\documentclass[journal]{vgtc}                          

\usepackage{bbm}
\usepackage{bbold}
\usepackage{amsmath}
\usepackage{algorithm}
\usepackage{amsfonts}
\usepackage{paralist}
\usepackage{tabularx, ragged2e}
\usepackage[noend]{algpseudocode}
\usepackage{color}

\usepackage{graphicx}
\usepackage{times}
\usepackage{enumitem}

\usepackage{CJKutf8}
\usepackage{wrapfig}

\title{Revisiting the Modifiable Areal Unit Problem in Deep Traffic Prediction with Visual Analytics}

\author{
Wei Zeng, Chengqiao Lin, Juncong Lin, Jincheng Jiang, Jiazhi Xia, Cagatay Turkay, Wei Chen
}

\authorfooter{
\item
Wei Zeng and Jincheng Jiang are with Shenzhen Institutes of Advanced Technology, Chinese Academy of Sciences, China. E-mail: \{wei.zeng, jc.jiang\}@siat.ac.cn.

\item
Chengqiao Lin and Juncong Lin are with Xiamen University, China. E-mail: \{linchengqiao, jclin\}@xmu.edu.cn. Juncong Lin is the corresponding author.

\item
Jiazhi Xia is with Central South University, China. E-mail: xiajiazhi@csu.edu.cn.

\item
Cagatay Turkay is with University of Warwick, UK. E-mail: cagatay.turkay@warwick.ac.uk.

\item
Wei Chen is with the State Key Lab of CAD\&CG, Zhejiang University, China. E-mail: chenwei@cad.zju.edu.cn.
}

\onlineid{1045}

\vgtccategory{Research}
\vgtcpapertype{Application/design study}

\vgtcinsertpkg

\newcommand{\red}[1]{{\color{black}{#1}}}

\algnewcommand\algorithmicforeach{\textbf{for each}}
\algdef{S}[FOR]{ForEach}[1]{\algorithmicforeach\ #1\ \algorithmicdo}

\teaser{
  \centering
  \includegraphics[width=0.925\linewidth]{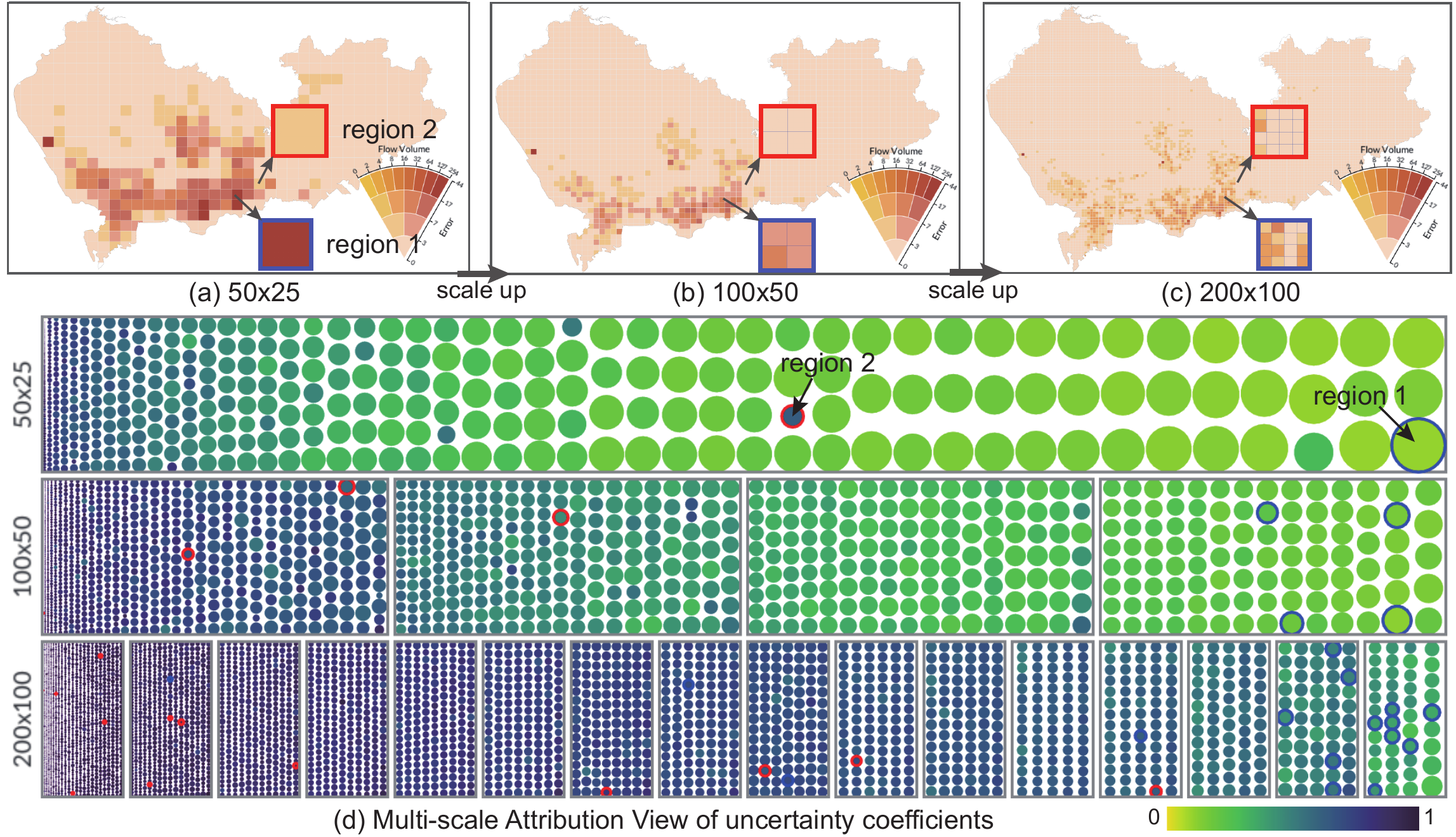}
  \vspace{-4mm}
  \caption{
  Diagnosing deep traffic predictions across multiple scales.
  We design \emph{Bivariate Maps} (a-c) to depict traffic volumes and prediction errors simultaneously across space, and \emph{Multi-scale Attribution View} (d) to compare scale-independent metrics across scales.
  One interesting observation here is that the volume of region 2 is mainly coming from a particular sub-region on the western end of the region at scale 200$\times$100.
  This discrepancy is also highlighted in the 50$\times$25 attribution view where region 2 shows a higher level of uncertainty as can be seen by the blue coloured dot.
  }
  \label{fig:teaser}
}

\abstract{
Deep learning methods are being increasingly used for urban traffic prediction where spatiotemporal traffic data is aggregated into sequentially organized matrices that are then fed into convolution-based residual neural networks. However, the widely known modifiable areal unit problem within such aggregation processes can lead to perturbations in the network inputs. This issue can significantly destabilize the feature embeddings and the predictions -- rendering deep networks much less useful for the experts. This paper approaches this challenge by leveraging unit visualization techniques that enable the investigation of many-to-many relationships between dynamically varied multi-scalar aggregations of urban traffic data and neural network predictions. Through regular exchanges with a domain expert, we design and develop a visual analytics solution that integrates 
1) a \emph{Bivariate Map} equipped with an advanced bivariate colormap to simultaneously depict input traffic and prediction errors across space,
2) a \emph{Moran's I Scatterplot} that provides local indicators of spatial association analysis, 
and 3) a \emph{Multi-scale Attribution View} that arranges non-linear dot plots in a tree layout to promote model analysis and comparison across scales. We evaluate our approach through a series of case studies involving a real-world dataset of Shenzhen taxi trips, and through interviews with domain experts. We observe that geographical scale variations have important impact on prediction performances, and interactive visual exploration of dynamically varying inputs and outputs benefit experts in the development of deep traffic prediction models.
}

\CCScatlist{ 
 \CCScat{I.3.8}{Computer Graphics}{Application}{Geographical Visualization};
 \CCScat{H.5.2}{Information and Interfaces and Presentation}{User Interfaces}{Evaluation/Methodology}
}

\begin{document}

\firstsection{Introduction}

\maketitle

Traffic prediction is a key tool for urban transportation and urban planning helping analysts and planners in improving traffic management and control~\cite{wang2012understanding}.
As a result, numerous traffic prediction algorithms have been developed within the last few decades, such as the auto regressive integrated moving average (ARIMA)~\cite{moorthy_1988_short, willianms_1998_urban} that takes advantage of repeating occurrences in temporal historical data.
However, such conventional methods are usually limited when it comes to modeling the complex non-linear spatial and temporal properties of urban traffic.
Recently, deep neural networks (DNNs) showed superior performance in traffic prediction.
DNN approaches (\emph{e.g.},~\cite{zhang_2017_deep, yao_2018_deep}) typically partition an underlying territory into grids, aggregate in- and out-flows in each grid, and model the spatio-temporal flows as a sequence of raster images.
In this way, urban traffic can be modeled and predicted using a convolution-based residual neural network~\cite{he_2016_deep}.

The flow aggregation step is highly crucial in this aforementioned process, which however, is subject to the modifiable areal unit problem (MAUP)~\cite{gehike_1934_certain, openshaw_1984_modifiable}: aggregations are influenced by both partition shapes (\emph{e.g.}, grids \emph{vs.} administrative units) and scales (\emph{e.g.}, coarse \emph{vs.} fine granularity) of the spatial partition units (see Fig.~\ref{fig:maup}).
The differences in flow aggregations, which are consumed as network inputs, can cause significant distortions to network outputs, since DNNs generally suffer from adversarial perturbation problem~\cite{Moosavi_2017_universal, zheng_2016_improving}.
This is counter-productive for transportation experts who expect stable and reliable outputs from predictive models~\cite{xu_2014_accurate}.
Therefore, it is critical to incorporate approaches that consider the impact of MAUP within the diagnosis of deep traffic predictions. 

This work seeks to address this need with a visual analytics approach.
This is nevertheless a non-trivial task.
First, urban traffic exhibits dynamic spatial variances.
Domain experts would like to analyze spatial associations of prediction accuracies \emph{vs.} localized traffic aggregations.
However, conventional side-by-side choropleth maps have limitations in presenting such information simultaneously in a way that effectively supports comparison~\cite{pena2020comparison}.
Second, existing methods~\cite{zhang_2017_deep, yao_2018_deep} measure prediction accuracy using a single numeric statistic, \emph{i.e.}, root mean square error (RMSE), which neglects the uniqueness of individual region and is not comparable over scales. Considering how critical it is to understand how perturbations in inputs affect the outputs in improving machine learning models~\cite{wexler_2019_what-if}, effective model building requires methods that can support the exploration of individual regions in relation to scale-independent metrics~\cite{duque_2018_s-maup}.

To address these challenges, we present a visual analytics solution with three main visualization modules:
(i) A \emph{Bivariate Map} encodes traffic volumes and prediction errors simultaneously on a bivariate map.
Here we employ a value-suppressing uncertainty palette (VSUP)~\cite{michael_2018_value-suppressing} to encourage more cautious inspections of error-high regions.
(ii) A \emph{Moran's I Scatterplot} depicts local indicators of spatial association (LISA)~\cite{anselin_1995_lisa} indices of urban traffic at each region and at a local tract \red{that corresponds to the size of convolution matrix adopted by the DNN model}.
(iii) A \emph{Multi-scale Attribution View} adopts the idea of nonlinear dot plot~\cite{rodrigues_2018_nonlinear} to encode each region as a dot.
All modules leverage unit visualization techniques, \emph{i.e.}, visualizations where each visual element correspond to a single data point rather than depicting an aggregate~\cite{park_2018_atom}, in order to enable the investigation of models over individual data points.
We evaluate our system on a real-world dataset of Shenzhen taxi movements, and demonstrate its effectiveness through case studies and interviews with domain experts.

\vspace{1.5mm}
The main contributions of this work include:

\begin{itemize}

\vspace{-1.5mm}
\item
A visual analytics system that incorporates various unit visualization techniques, including bivariate choropleth map, scatterplot, and nonlinear dot plot, for diagnosing impacts of the MAUP on deep traffic prediction.

\vspace{-1.5mm}
\item
A new layout strategy for nonlinear dot plot that positions all dots in a compact manner, and supports flexible arrangement of multiple dot plots in a tree layout.

\vspace{-1.5mm}
\item
Illuminating insights revealed from case studies, such as impacts of spatial variations on prediction accuracy, which provides promising directions for improving deep traffic prediction. 

\end{itemize}

\section{Related Work}
\label{sec:related_work}

\vspace{1mm}
\noindent
\textbf{Geographical Partition}:
Partitioning geographical space into appropriate regions is a necessary step in many applications, \emph{e.g.}, urban form studies~\cite{shen_2017_streetvizor} and movement visualization~\cite{zeng_2013_visualizing}.
In its simplest form, a territory can be partitioned into equal-sized grids, yielding a $w \times h$ matrix.
Many prior studies adopted this approach when visualizing movement data, \emph{e.g.},~\cite{weng_2019_srvis, deng_2020_airvis}.
Besides grid-based division, another popular partition method is by administrative units, which are typically in irregular shapes.
Examples include municipalities, districts, and census tracts.
In many circumstances, it is also desirable to construct multiple scales of partitions, which are internally homogeneous and occupy contiguous regions in space.

The two dimensional properties of partition units, \emph{i.e.}, \emph{shaping} effect referring to changes in the shape, and \emph{scalar} effect referring to changes in the size, can cause the MAUP, which states that aggregations by different partitions may present different (or even wrong) patterns~\cite{gehike_1934_certain, openshaw_1984_modifiable}.
Recent works of utilizing DNNs for traffic prediction~\cite{zhang_2017_deep, yao_2018_deep} partitioned the space by grids, generating fixed-size matrices for network consumption.
Nevertheless, effects of the MAUP on network inputs and prediction performances were neglected.
This work supplements the gap with a visual analytics approach.

\vspace{1mm}
\noindent
\textbf{MAUP Analysis and Visualization}:
Understanding shaping and scalar effects of the MAUP requires a spatial analytical approach.
On one hand, attributes of spatial data exhibit similarities in nearby spatial units, as stated by the first law of geography $-$ ``everything is related to everything else, but near things are more related than distant things''; 
on the other hand, experiments also revealed that values for a particular measure vary across all spatial units~\cite{brunsdon_1996_spatial_nonstationarity}. 
These characteristics derive the development of spatial autocorrelation analysis, including Geary's C~\cite{geary_1954_contiguity}, Moran's I~\cite{moran_I}, etc.
In addition to numerical indicators, researchers also exploited exploratory data analysis approaches to investigate the effects of the MAUP.
Among them, attribute signatures~\cite{turkay_2014_attribute} utilized small multiple charts to provide visual summaries of different statistical variables under the varying levels of aggregation.
Goodwin et al.~\cite{goodwin_2016_visualizing} modeled MAUP as a geo-visual parameter space analysis~\cite{sedlmair_2014_visual} problem, and designed a set of glyphs to encode correlations of a given variable at different scales.
Zhang et al.~\cite{zhang_2016_visual} coupled spatial lens with glyph-based designs to maintain the context of the spatial clusters at different spatial scales.

Those studies focused on understanding the MAUP across an entire or a subset of dataset, while some other exploratory data analytics explored the MAUP by examining individual data points.
Nelson and Brewer~\cite{nelson_2017_evaluating} depicted correlations between a variable and itself across multiple scales.
Zhang et al.~\cite{zhang_2016_visualizing} interactively explored and compared the impact of geographical variations for multivariate clustering.
Huang et al.~\cite{huang_2019_exploring} recently explored choropleth classification results upon data uncertainty.
These works in like manner adopt choropleth maps and scatterplots to depict spatial information and numerical variables, respectively.

Our work enables investigation of individual regions.
A distinguishing feature of our system is that we design \emph{Multi-scale Attribution View}, allowing users to compare changes caused by the scalar effect and corresponding model performances across multiple scales.

\vspace{1mm}
\noindent
\textbf{Visual Interpretation for Deep Learning}:
Deep learning (DL) has advanced fields such as image and natural language processing.
However, DL models are regarded as a `black box', hindering their usability in many domains such as traffic prediction, which requires stable prediction outputs.
An example for stability issues is the adversarial perturbation problem, \emph{i.e.}, marginal input differences may cause significant effects on the output~\cite{Moosavi_2017_universal, zheng_2016_improving}.
Cao et al.~\cite{cao_2020_analyzing} analyzed the robustness of DL models against adversarial examples by depicting the internal datapaths of how adversarial and normal examples diverge and merge in the prediction process.
Many visualizations have been developed to unveil internal working mechanism, especially the hidden layer behaviors, of DL models.
Examples for convolutional neural networks (CNNs) include~\cite{zeiler_2014_visualizing, liu_2017_towards, pezzotti_2018_deepeye, lapuschkin_2019_unmasking}, and recurrent neural networks (RNNs) include~\cite{ming_2017_understanding, strobelt_2018_lstmvis, kwon_2019_retainvis, shen_2020_visual}.
Interested readers are referred to~\cite{Liu2017, hohman_2019_visual, yuan2021survey} for comprehensive surveys.

Other than depicting internals of DL models, some other methods focus on understanding the relationships between input features and output predictions.
These methods usually assign each feature an importance score to indicate how it impacts the final prediction.
For instance, SHAP~\cite{lundberg2017unified}$-$Shapley values from game theory$-$measured feature attribution from a local perspective, and calculated the feature contribution of a data point by comparing it with a set of reference data points.
The method is more consistent and locally accurate, in comparison with global solutions.
A recent work $-$ the What-If Tool~\cite{wexler_2019_what-if} $-$ supports both analysis of decision on a single data point, and understanding of model behavior across an entire dataset.
Users are allowed to explore how general changes to data points affect predictions.
An important feature of the tool is flexible sorting, which is crucial for user-centric explainable AI~\cite{wang_2019_designing}.

Specifically, this work seeks to understand the relationships between network input of traffic aggregations that are affected by the MAUP, and output predictions suffering from the adversarial perturbation problem.
We leverage various unit visualization techniques to enable unit-level investigation.
\section{Background, Task, and System Overview}
\label{sec:overview}

\begin{figure}
  \centering
  \includegraphics[width=0.45\textwidth]{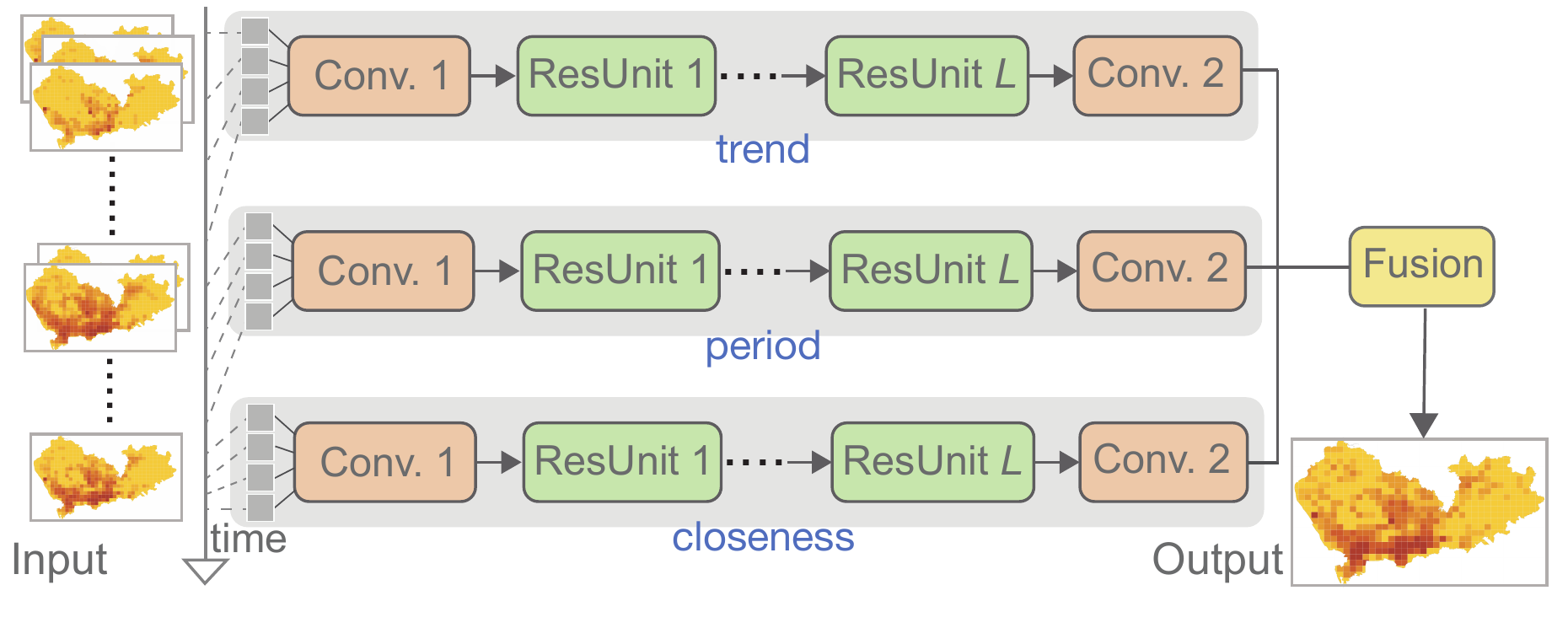}
  \vspace{-5mm}
  \caption{Simplified architecture of ST-ResNet adopted in our work.}
  \vspace{-4mm}
  \label{fig:st-net}
\end{figure}

\begin{table}[!htb]
    \centering
    \vspace{-2mm}
    \caption{Meanings of all notations.} 
    \vspace{-2mm}
    \resizebox{0.9\linewidth}{!}{
        \begin{tabular}{|c|l|} \hline
        \textbf{Notation} & \textbf{Description}  \\ \hline\hline
        $\mathcal{T}; t$ & Set of all time slots; a time slot. \\ \hline
        $\mathcal{M}; m$ & Set of all movements; a movement. \\ \hline
        $\mathcal{R}; r$ & Set of all regions at a partition shape and scale; a region. \\ \hline
        $\mathcal{G}; g$ & Set of all grids for network input; a grid. \\ \hline
        $x_{r,t}$ or $x_{g,t}$ & Aggregated traffic in time slot $t$ and region $r$ or grid $g$. \\ \hline
        $y_{r,t}$ or $y_{g,t}$ & Predicted traffic in time slot $t$ and region $r$ or grid $g$. \\ \hline
        \end{tabular}
    }
  \vspace{-3mm}
    \label{table:notations}
\end{table}

The section introduces the research background (Sec.~\ref{ssec:bg}), followed by the analytical tasks (Sec.~\ref{ssec:tasks}) and system overview (Sec.~\ref{ssec:overview}).
To facilitate the discussion, we list down common notations adopted in this work as in Table~\ref{table:notations}.

\subsection{Background}
\label{ssec:bg}
The interest in the research efforts on traffic prediction have recently shifted towards DNNs.
ST-ResNet~\cite{zhang_2017_deep} as a pioneering work, models urban traffic as temporal-varying matrices, and employs ResNet~\cite{he_2016_deep} to encapsulate the spatio-temporal dynamics.
Fig.~\ref{fig:st-net} presents a simplified architecture of ST-ResNet adopted in this work.
As a first step, the method partitions the entire area into non-overlapping grids at time slot $t$, and aggregates traffic in each grid $g$ as $x_{g,t}$.
In this way, a flattened matrix $X_t \in \mathbb{R}^{w\times h}$ representing aggregated traffic in all grids of size $w \times h$ is constructed. 
Next, ST-ResNet consumes a series of $\{X_{t} | t \in \{t_0,\cdots,t_n\}\}$ as network input, and learns the periodic patterns in the historical traffic.
Specifically, ST-ResNet models the temporal dependency as:
i) \emph{trend} for weekly trend, ii) \emph{period} for daily periodicity, and iii) \emph{closeness} for recent time dependence.
Finally, ST-RestNet fuses the three learned periodic patterns, and produces a matrix $Y_{t_{n+1}} \in \mathbb{R}^{w\times h}$ as traffic prediction for time slot $t_{n+1}$.

\begin{figure}
  \centering
  \includegraphics[width=0.45\textwidth]{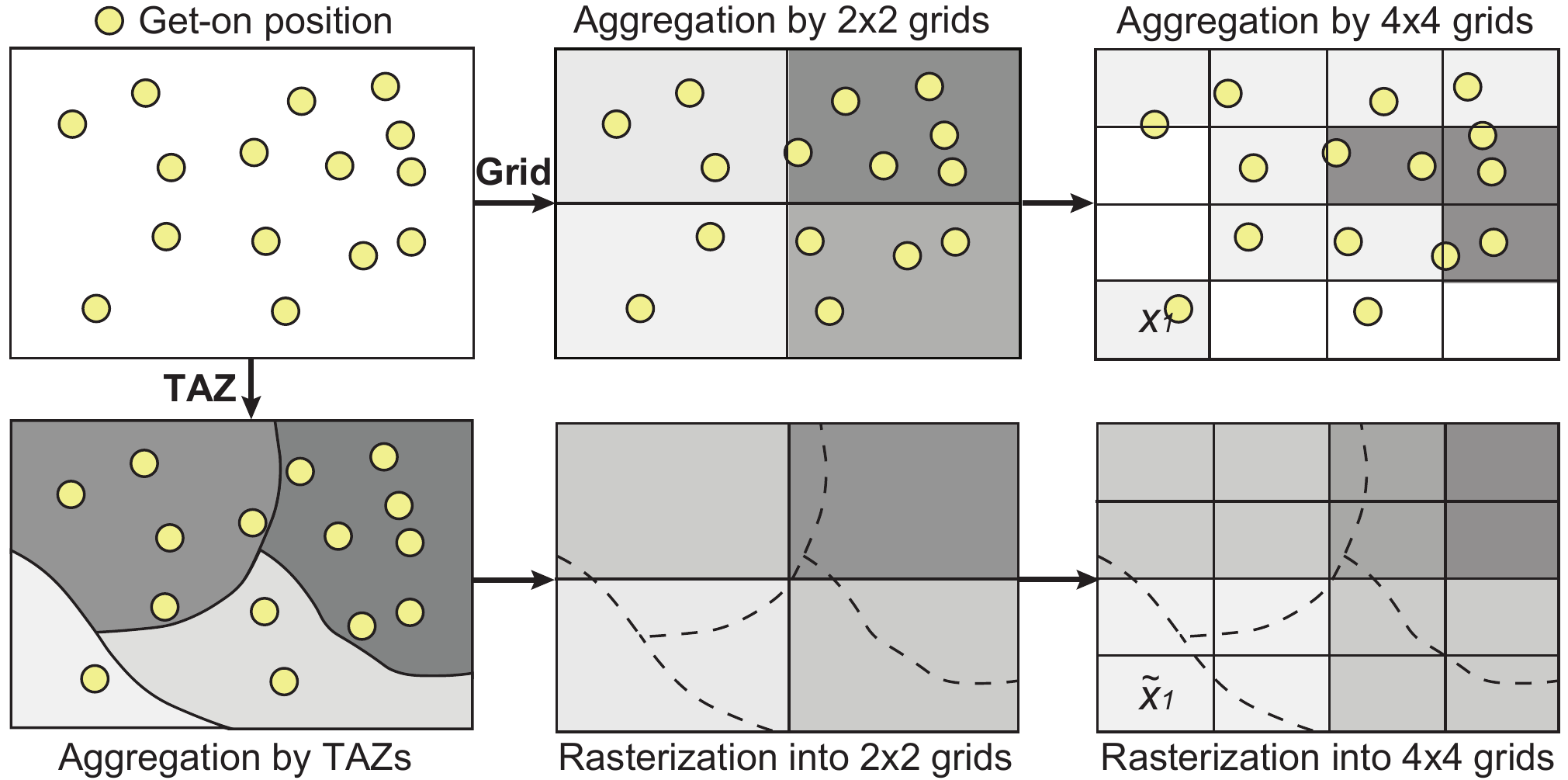}
  \vspace{-4mm}
  \caption{Aggregations of urban traffic are subject to shapes (\emph{e.g.}, grid \emph{vs.} TAZ) and scales (\emph{e.g.}, 2$\times$2 \emph{vs.} 4$\times$4) of partition units.}
  \vspace{-4mm} 
  \label{fig:maup}
\end{figure}

However, traffic aggregations are subject to shapes and scales of the spatial partition units, \emph{i.e.}, the MAUP~\cite{gehike_1934_certain, openshaw_1984_modifiable}.
As shown in Fig.~\ref{fig:maup}, we can aggregate urban traffic by different shapes, such as grids (top) or traffic analysis zones (TAZs) (bottom), and by different scales, such as $2\times2$ (middle) or $4\times4$ (right) grids.
Notice that TAZs are in irregular shapes, we need to further rasterize the results into grids to feed the neural network.
For example, we can rasterize the TAZs into $4\times4$ grids, as the same size with $4\times4$ Grid partition.
Though grid sizes are the same, grid values by griding and TAZ partitions are different, \emph{e.g.}, $x_1 \neq \tilde{x}_1$.
The differences may be marginal, but it may cause significant effects on the output, because DNNs generally suffer from adversarial perturbation problem~\cite{Moosavi_2017_universal, zheng_2016_improving}.

In the past eight months, we closely worked with a collaborating researcher (\emph{CR}) specialized in the field of geography and traffic analysis.
\emph{CR} is interested in applying deep learning techniques in traffic prediction.
In the beginning, we divided the studying area into $50 \times 25$ grids that is close to the setting adopted in~\cite{zhang_2017_deep}, and employed ST-ResNet for traffic prediction.
We showed \emph{CR} the prediction results in terms of \emph{RMSE}, which is a typical metric for evaluating prediction accuracy.
However, \emph{CR} questioned the choice for $50 \times 25$ grids, and introduced the MAUP to us.
\emph{CR} expected a visual analytics to diagnose the predictions upon different partition shapes and scales.

\subsection{Analytical Tasks}
\label{ssec:tasks}
To better understand the problem domain, we conducted several rounds of \red{semi-structured} interviews with \red{\emph{CR}}.
Rather than attempting to examine the internal mechanisms of ST-ResNet, \emph{CR} is more interested in investigating correlations between input features and output predictions, such that he can manipulate data processing to fine-tune the results.
In consultation with \emph{CR}, we distilled three research goals:
\emph{G1}: understand the MAUP effect on traffic aggregations;
\emph{G2}: understand the output predictions upon variances in input features;
and \emph{G3}: support the exploration of an individual region.
To this end, we compile a set of analytical tasks:

\begin{enumerate}[label={T.\arabic*:}]
\vspace{-2mm}
\item
\textbf{Spatial Variation Exploration}.
Urban traffic exhibit dynamic spatial variance.
The experts would like to explore traffic distributions upon shaping and scaling effects over space (G1 \& G3), and how the output predictions vary accordingly (G2 \& G3).
This task requires the solution to present spatially varying bivariate variables simultaneously.

\vspace{-2mm}
\item
\textbf{Spatial Association Analysis}.
Furthermore, ST-ResNet applies 2D convolutional operations on the flattened matrix of traffic aggregations.
A 2D convolution applies an element-wise multiplication of neighborhood matrix indices and a small matrix of weights, and sums up the results into a single cell.
Hence, it is necessary to support the exploration of spatial associations among input traffic at individual regions and at local tracts \red{that are the surrounding regions of a region under investigation }(G1), and explore the effects on output predictions (G2).

\vspace{-2mm}
\item
\textbf{Scale-independent Comparison}.
Diagnosing the scaling effect \red{needs} to consider feature variances upon scales.
The comparison shall remove the scaling effect of different attribution ranges.
Besides, the criteria are measured upon each region, rather than a summary statistic on the entire area (G3).
The visual analytics should incorporate unit visualization that provides flexibility for investigating a single data point.
\end{enumerate}

\subsection{System Overview}
\label{ssec:overview}
The system mainly consists of three modules: 1) \textit{data preprocessing}, 2) \textit{prediction \& analysis}, and 3) \textit{interactive visualization}.
In \textit{data preprocessing} stage, we process the raw data of two-month (59 days) taxi movements into network consumable matrices (Sec.~\ref{ssec:preprocessing}).
We select two types of partition \emph{shapes}, \emph{i.e.}, grid and TAZ; and three levels of \emph{scales}, \emph{i.e.}, 50$\times$25, 100$\times$50, and 200$\times$100.
We also experimented with scale 400$\times$200, but the network failed to converge, probably because there are too many zero values in the input matrix. 
For TAZ shapes, the scales refer to matrix size after rasterization.
After preprocessing, we generate six sequences (two \emph{shapes} $\times$ three \emph{scales}) of flattened matrices representing the two-month taxi movements.

In \textit{prediction \& analysis} stage, we select the flattened matrices for the first 52-days from each matrix sequence as training data, generating in total six ST-ResNet models.
Each model is used to predict traffic for the remaining 7-day testing data.
Last, we evaluate the prediction accuracy using scale-independent metrics (Sec.~\ref{ssec:statistics}).
Both \emph{data preprocessing} and \textit{prediction \& analysis} stages are conducted offline on a workstation with 8 core 3.2 GHz AMD Ryzen 7 2700 CPU and a NVIDIA GeForce RTX 2080Ti graphics card.
The training takes about 20 hours for the scale 50$\times$25, and up to two days for the scale 200$\times$100.

The processed matrices and analysis results are passed to the \textit{interactive visualization} module.
The interface mainly integrates three coordinated views of a \emph{Bivariate Map}, a \emph{Moran's I Scatterplot}, and a \emph{Multi-scale Attribution View} (Sec.~\ref{sec:visual-design}).
The interface is implemented in LWJGL, with the map, scatterplot, and dot plot rendered with OpenGL, and overlaid buttons and text realized in NanoVG.
The system currently runs on an Intel Core i7 2 2.8GHz MacBook Pro with 16GB memory and an AMD Radeon R9 M370X graphics board.

\section{Data Processing and Model Evaluation}
\label{sec:model}

\subsection{Input Dataset}
\label{ssec:raw}
The input dataset consists of the following data in Shenzhen, China.

\vspace{1.5mm}
\noindent
\textbf{Taxi Transaction Records}:
The data record taxi transactions made by over 20k taxis during the period from 1 Jan. 2019 to 28 Feb. 2019 (59 days).
There are about 800k transactions recorded per day, summing up to over 47 million transactions in total.
Each taxi transaction is regarded as an individual movement.
For each movement $m$, the following attributes are recorded: \emph{taxi ID}, \emph{price}, \emph{operating mileage}, \emph{get-on position} (denoted as $m_{p0}$) and \emph{time} ($m_{t0}$), and \emph{get-off position} ($m_{p1}$) and \emph{time} ($m_{t1}$).
The raw data contains various corrupt or inaccurate records, such as locations outside Shenzhen, or missing get-on/-off times, etc.
We cleaned up the data to alleviate the effects on the following experiments.
After data cleaning, there remain about 45 million valid transaction records.

\begin{figure}
  \centering
  \includegraphics[width=0.45\textwidth]{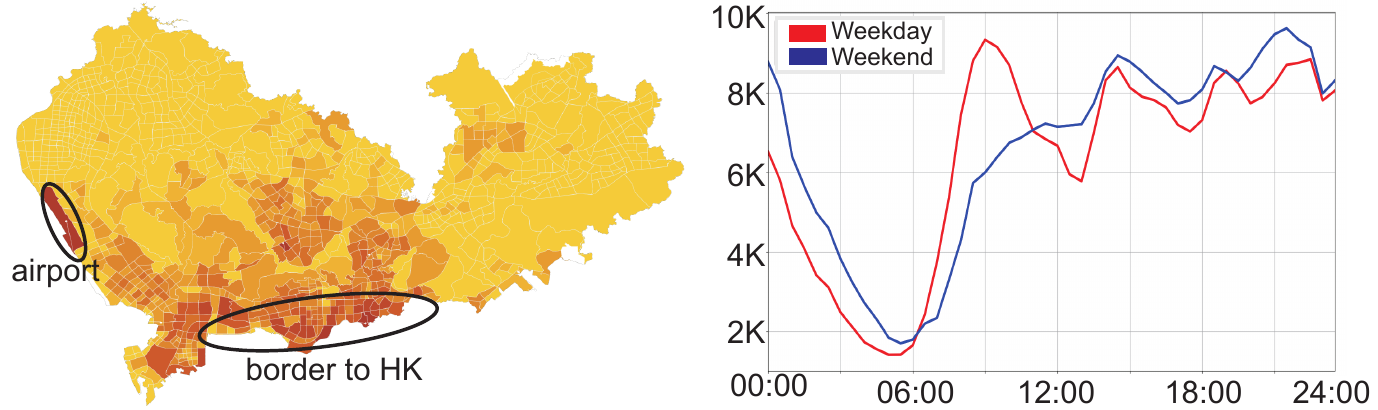}
  \vspace{-4mm}
  \caption{Distribution of taxi movements in space (left) and time (right).}
  \vspace{-4mm}
  \label{fig:Distribution}
\end{figure}

\vspace{1.5mm}
\noindent
\textbf{Traffic Analysis Zones}:
A TAZ is a geography unit constructed by census block information for tabulating traffic-related data.
The spatial extent of TAZs varies, which are typically large areas in the exurb and small blocks in central business districts.
In this way, the number of people in each zone is balanced, such that to better couple with conventional traffic planning and demand analysis.
This work leverages TAZs of 1066 zones delineated by Shenzhen transportation officials.
An illustration of the TAZs is presented in Fig.~\ref{fig:Distribution} (left).
Notice that the zones are typically small where traffic volumes are high.

\vspace{1.5mm}
Fig.~\ref{fig:Distribution} presents spatial (left) and temporal (right) distributions of journey origin locations and times averaged over every 30 minutes.
Fig.~\ref{fig:Distribution} (left) depicts that the taxi movements are concentrated in southern parts of the city, which are central business districts on border with Hong Kong.
There is also a high-volume zone in the west where the airport is located.
Less taxi movements are found in other places.
Fig.~\ref{fig:Distribution} (right) show averaged traffic on weekdays (red) and weekends (blue), respectively.
The distributions show dramatic drops of taxi movements after midnight, whilst the drop delays about one hour on weekends.
A morning peak hour can be found at around 9:00 on weekdays.
Some other peaks can be found at around 15:00, 18:00, and 22:00 on both weekdays and weekends.
The dramatic spatial and temporal variances bring challenges for traffic predictions.




\subsection{Data Preprocessing}
\label{ssec:preprocessing}
ST-ResNet is designed to consume a sequence of fixed-sized matrices.
Before training, we need to process the raw taxi movements to fit the network inputs.
The preprocessing takes the following steps:

\begin{itemize}

\vspace{-2mm}
\item
\textbf{Geographical partition:}
To explore the \emph{shaping} effect, two partition shapes are tested: grids and TAZs.
The studying area has been divided into 1066 partitions in TAZ, yet we need to determine the size for Grid partition. 
We check ranges of the studying area, which are [113.775, 114.629] in longitude, and [22.443, 22.855] in latitude.
That is, the range of longitude is roughly two times of latitude.
To accommodate the matrix size adopted in ST-ResNet~\cite{zhang_2017_deep}, that is 32$\times$32, we choose a similar size of 50$\times$25 as the coarsest scale.
Next, to diagnose the \emph{scalar} effect, we would like to compare multiple partition scales.
\emph{CR} noted that smaller matrix size will make each grid cover a too large underlying area, which is useless for traffic prediction.
Therefore we choose finer scales of 100$\times$50 and 200$\times$100.

\vspace{-1mm}
\item
\textbf{Aggregation:}
We split each day into 48 time slots, with each slot covering 30 minutes.
There are in total $2832$ ($59\times48$) time slots, \emph{i.e.}, $|\mathcal{T}| = 2832$.
Next, we assign taxi movements into each time slot $t$ based on their getting-on times, which can be represented as $\mathcal{M}_t = \{ m \in \mathcal{M} | m_{t0} \in t\}$.
Next, we separate $\mathcal{M}_t$ into regions based on their get-on positions.
The regions of a particular partition shape and scale are denoted as $\mathcal{R} =\{r_{i}\}_{i=1}^n$, where $n = 1066$ for TAZ partition, and $n = 1250$ for scale $50\times25$, $n=5000$ for scale $100\times50$, and $n=20000$ for scale $200\times100$ under Grid partition.
The regions are non-overlapping and fill up the studying area.
Thus, we can find unique region $r$ for a movement $m$ that meets the condition $m_{p0} \in r$.

In this way, we derive a nonnegative integer $x_{r,t}$ counting the number of movements in a region $r$ and time slot $t$.


\end{itemize}

Aggregations for Grid partition are naturally in a matrix format that can be directly consumed by ST-ResNet.
On the other hand, TAZ partitions are in arbitrary shapes, and possess only one scale.
In order to be used as network inputs and enable multi-scale comparison, we further adopt the following steps for TAZ partitions.
\vspace{-1mm}
\begin{itemize}
\item
\textbf{Rasterization:}
The process is to convert TAZ-based traffic aggregations $X^r_{t}$ into a raster matrix $X^g_{t} \in \mathbb{R}^{w\times h}$.
Each grid $g$ could intersect with arbitrary number of TAZ regions $\{r_i\}_{i=1}^k$.
We calculate the value for each grid $x_{g,t}$ as:
\vspace{-3mm}
\begin{equation}
x_{g,t} = \sum_{i=1}^k x_{r_i,t} \times \frac{S(r_i \cap g)}{S(r_i)}
\label{equ:raster}
\end{equation} \vspace{-4mm}

where $S(\cdot)$ stands for the area of a region, and $r_i \cap g$ indicates the intersection between $r_i$ and $g$.






\end{itemize}

By this, we generate a sequence of raster matrices $\{X_t | X_t\in\mathbb{R}^{w \times h}\}$ as network inputs for both Grid and TAZ partitions.
\red{Note that the rasterization of TAZ partition is employed to fit in ST-ResNet inputs only.
One can easily compute predicted flow volume for each TAZ partition from predicted flow volumes of grid partitions by ST-ResNet.
Nevertheless, the rasterization operation obliterates neighborhood relationships among original TAZ partitions, which may cause negative effects on traffic predictions.
This is regarded as a limitation of deep traffic prediction with 2D convolutions~\cite{yao_2018_deep}.}

\subsection{Scale-Independent Evaluation Metrics}
\label{ssec:statistics}
To support multi-scale comparison (\textbf{T.3}), the comparison metrics should be scale-independent to remove the scaling effect of different value ranges.
We select three metrics satisfying this requirement~\cite{duque_2018_s-maup}: percentage RMSE (\emph{PRMSE}), uncertainty coefficient (\emph{U}), and correlation coefficient (\emph{CORR}). 
The coefficients are useful for comparing different forecast models, \emph{e.g.}, whether a sophisticated model is, in fact, any better than a simple one that repeats the last observed value.
We measure the metrics for each grid, such that to support the goal of exploring a region (\emph{G3}).
This is possible because each grid $g$ possesses a sequence of predictions $\{y_{g,t}\}$ and observations $\{x_{g,t}\}$, for $t \in \{t_1,\cdots,t_n\}$.
The metrics can be measured as follows:

\begin{itemize}
\vspace{-2mm}
\item
\emph{PRMSE} measures variances between the predictions and observations, which is calculated as:

\vspace{-2.5mm}
\begin{equation}
PRMSE_g = \frac{1}{\overline{x}_{g}}\sqrt{\frac{1}{n}\sum\nolimits_{t=t_1}^{t_n}\left(y_{g,t} - x_{g,t}\right)^2}
\end{equation}

\vspace{-2.5mm}
where $\overline{x}_g$ is the mean value of observations in $g$.
Value of \emph{PRMSE} is positive, and is preferred to be close to 0.

\vspace{-2mm}
\item
Uncertainty coefficient ($U$) measures how well a time series of predictions match with a time series of observations.

\vspace{-1.5mm}
\begin{equation}
U_g = \frac{\sqrt{\frac{1}{n} \sum_{t=t_1}^{t_n}\left(y_{g,t}-x_{g,t}\right)^{2}}}{\sqrt{\frac{1}{n} \sum_{t=t_1}^{t_n} y_{g,t}^{2}}+\sqrt{\frac{1}{n} \sum_{t=t_1}^{t_n} x_{g,t}^{2}}}
\end{equation}

\vspace{-2.5mm}
Value of \emph{U} ranges from [0, 1], and is preferred to be close to 0.

\vspace{-2.5mm}
\item
Correlation coefficient (\emph{CORR)} measures how strong \red{is the} relationship between the predictions and observations.

\vspace{-4.5mm}
\begin{equation}
CORR_g= \frac{\sum_{t=t_{1}}^{t_{n}}\left(y_{g,t}-\overline{y}_g\right)\left(x_{g,t}-\overline{x}_g\right)}{\sqrt{\sum_{t=t_1}^{t_n}(x_{g,t} - \overline{x}_g)^2} \sqrt{\sum_{t=t_1}^{t_n}(y_{g,t} - \overline{y}_g)^2}}
\end{equation}

\vspace{-2.5mm}
Value of \emph{CORR} ranges in [-1, 1], where 0 indicates no relationship and +/-1 indicate perfect positive/negative correlations.

\end{itemize}

\section{Visualization Design}
\label{sec:visual-design}


To address the analytical tasks (Sec.~\ref{ssec:tasks}), we follow the following rationales in designing the interface:

\begin{enumerate}[label={R.\arabic*:}]

\vspace{-2mm}
\item \textbf{Coordinated Views:}
Various input features could cause prediction errors, \emph{e.g.}, spatial heterogeneity, local autocorrelation, etc.
To comprehensively reveal the correlations between input features and output predictions, coordinated multiple views (CMV) that support visual analytics from multiple joint-perspectives would fulfill the requirement.

\vspace{-2mm}
\item \textbf{Overview + Details:}
The visual analytics should provide an overview of data attributions over space and across multiple scales, and allow users to explore details on demand.
Efficient selection operations shall be incorporated to support examination of a single or a subset of data points.

\vspace{-2mm}
\item \textbf{Unit Visualization:}
Aggregated statistics and visualization would support the analytical tasks, such as to explore traffic variance over space (\textbf{T.1}).
Instead, unit visualization can maintain the identity of each visual mark and its relation to a data item~\cite{park_2018_atom}.
Unit visualization \red{needs} to also support sorting by particular criteria of importance, which is encouraged for user-centric explainable AI~\cite{wang_2019_designing}.
\end{enumerate}

Based on these rationales, we design a CMV system that primarily incorporates three unit visualization modules of a \emph{Bivariate Map} (Sec.~\ref{ssec:map_view}), a \emph{Moran's I Scatterplot} (Sec.~\ref{ssec:moran_plot}), and a \emph{Multi-scale Attribution View} (Sec.~\ref{ssec:unit_plot}).
We also integrate a set of interactions (Sec.~\ref{ssec:interactions}) to facilitate system exploration.


\begin{figure}[t]
  \centering
  \includegraphics[width=0.455\textwidth]{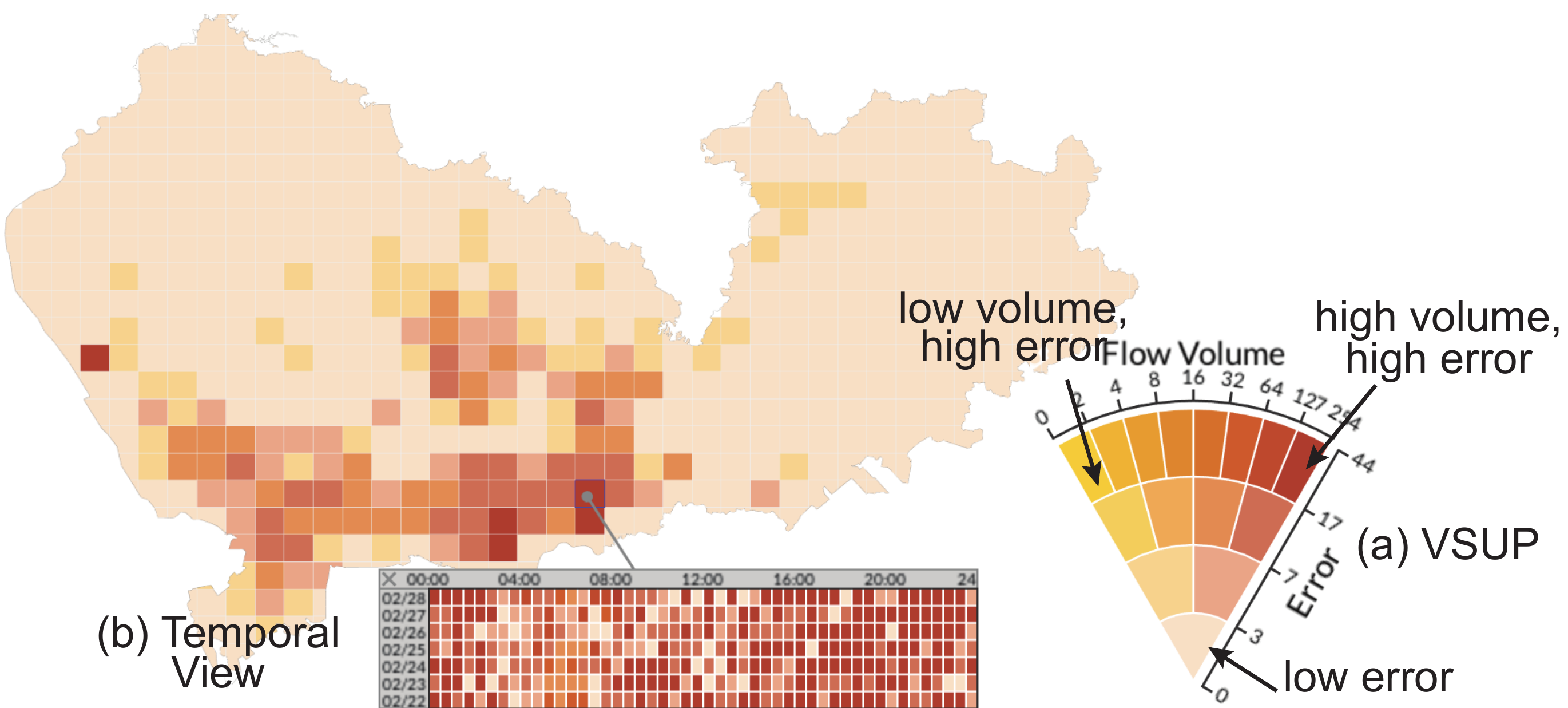}
  \vspace{-3mm}
  \caption{
  Bivariate Map adopts a value-suppressing uncertainty palette (VSUP) (a) to simultaneously present traffic volumes and prediction errors over space.
  Temporal View (b) is shown for a selected partition, presenting temporal variations over all time stamps.
  }
  \label{fig:bivariate_map}
\end{figure}

\begin{figure}[t]
  \centering
  \includegraphics[width=0.455\textwidth]{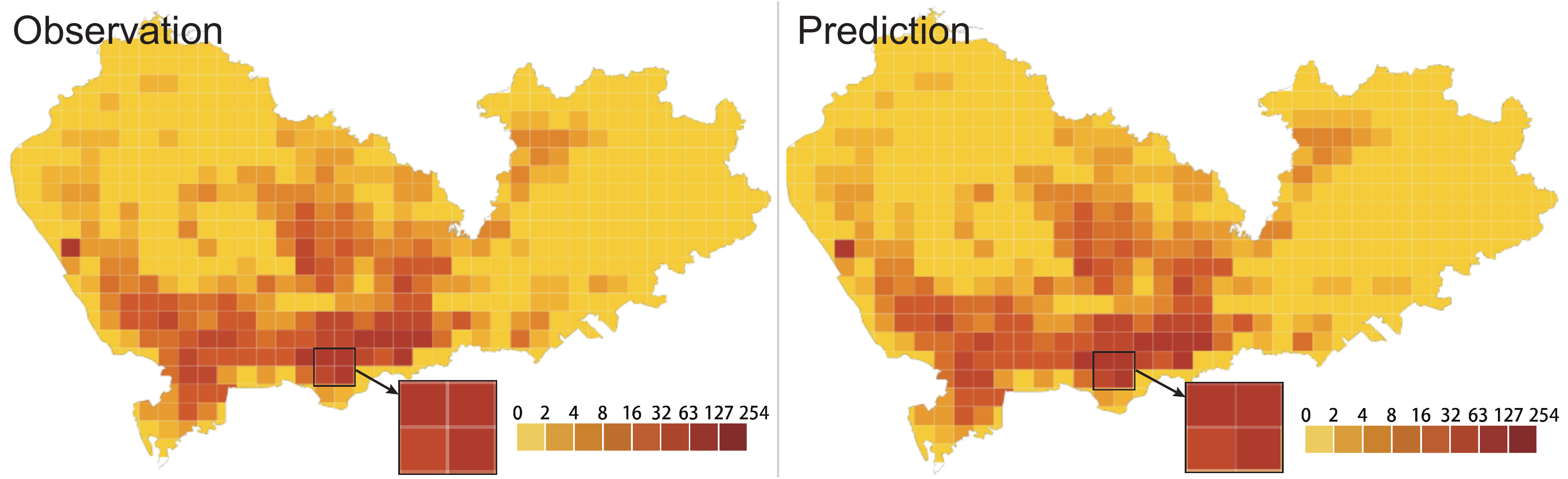}
  \vspace{-4mm}
  \caption{
  Arranging choropleth maps side-by-side is an alternative design choice for the bivariate map.
  }
  \vspace{-3mm}
  \label{fig:map_alter}
\end{figure}

\subsection{Bivariate Map}
\label{ssec:map_view}

We design Bivariate Map to support spatial variation exploration (\textbf{T.1}).
The view is essentially a bivariate choropleth map that simultaneously depicts traffic volume and prediction error at each grid.
The bivariate map is constructed as follows:
For each grid $g$, we compute the mean value of observed traffic $\overline{x}_g$ in the testing data as the first dimension, then compute the mean value of absolute prediction errors $\sum_{t=t_1}^{t_n}(|y_{g,t} - x_{g,t}|)/n$ as the second dimension.
We divide the first dimensional values into eight ranges, while the second dimensional values are divided into four ranges.
Then, we encode the two-dimensional values using a value-suppressing uncertainty palette (VSUP)~\cite{michael_2018_value-suppressing}.
As shown in Fig.~\ref{fig:bivariate_map}(a), the VSUP is in wedge shape, instead of a conventional bivariate colormap in square shape.
By this adaption, VSUP emphasizes those grids with higher prediction errors, \emph{i.e.}, colors towards outbound of the wedge.
From Fig.~\ref{fig:bivariate_map}, we can notice that grids with high prediction errors are mostly concentrated in the south, which also exhibit high traffic volumes.

The view allows users to select specific grids for in-depth investigation.
Upon selection, variations of traffic volumes and prediction errors over all testing time slots (7 days$\times$48 slots/day) are presented as a heatmap (Fig.~\ref{fig:bivariate_map}(b)).
The heatmap adopts the same color encodings as the VSUP.
Multiple grids can be selected for comparison, and heatmaps can be dragged around to mitigate occlusion.
Fig.~\ref{fig:bivariate_map}(b) presents temporal view of a grid in central business district, which produces the highest prediction errors under Grid partition at scale 50$\times$25.
We can observe that the grid exhibit high traffic volumes and high prediction errors throughout all the time slots.

\vspace{1mm}
\noindent
\textbf{Alternative design.}
Besides the bivariate map, an alternative design choice is to arrange two choropleth maps side-by-side.
An example is shown in Fig.~\ref{fig:map_alter}.
Here, the left one presents observed traffic volumes, while the right one presents predicted traffic volumes.
The views effectively depict dynamic spatial variations, and reveal strong correlations between observations and predictions.
However, prediction errors are not obvious, as grids on both sides show almost the same colors; see insets for an example.
Alternatively, we can also directly present prediction errors in the right-side choropleth map.
Though the design can better depict prediction errors, users need more efforts to link the grids in side-by-side views.  
Therefore, the design cannot compete with the bivariate map.

\begin{figure}[t]
  \centering
  \includegraphics[width=0.43\textwidth]{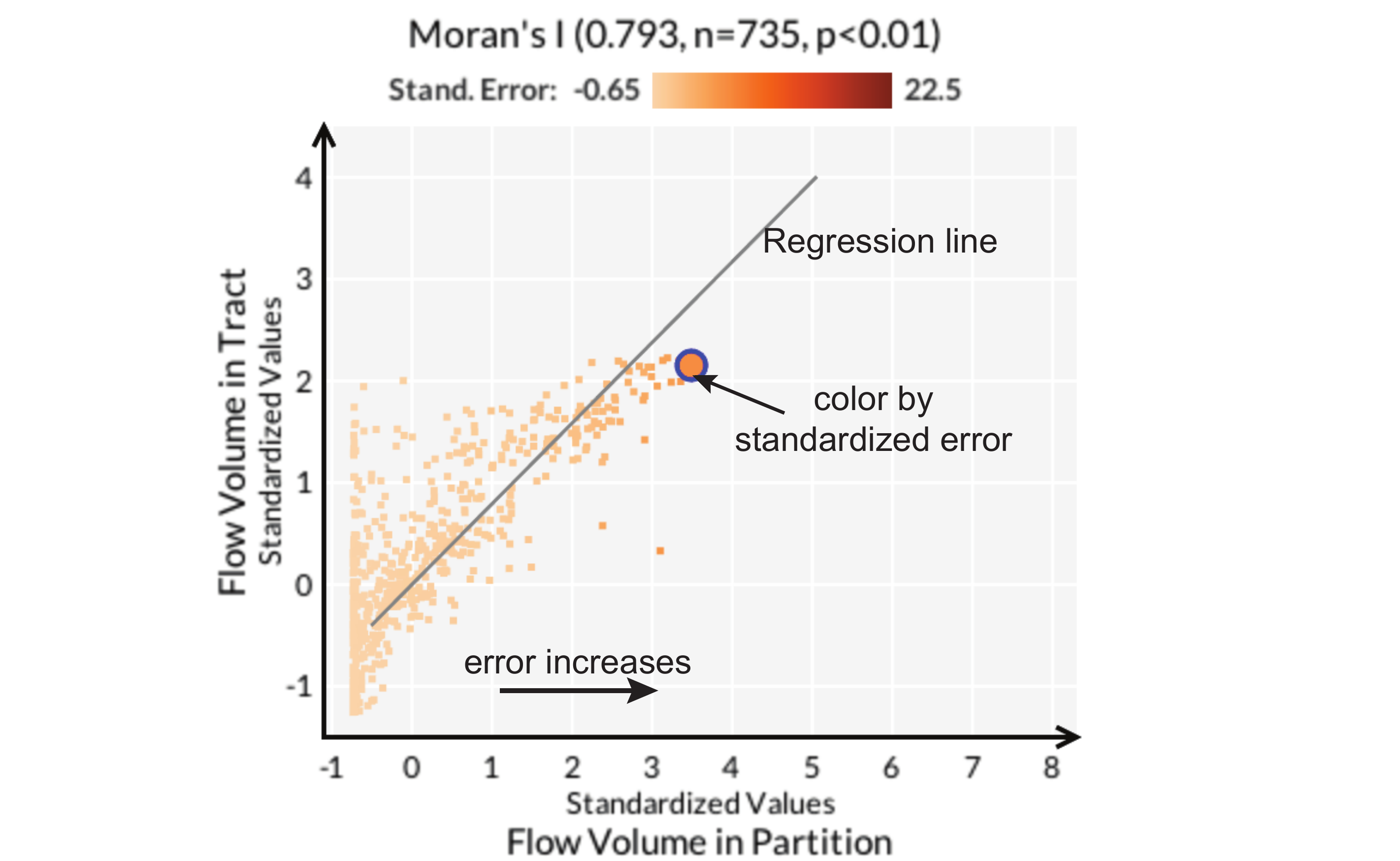}
  \vspace{-3mm}
  \caption{
  Moran's I Scatterplot depicts local autocorrelation of spatial association between traffic volume in regions and in local tracts.
  }
  \vspace{-5mm}
  \label{fig:moran}
\end{figure}
\subsection{Moran's I Scatterplot}
\label{ssec:moran_plot}
The Moran's I Scatterplot is designed to reveal the spatial autocorrelation of traffic volumes in each \red{grid-based} partition and local tract (\textbf{T.2}).
There exist many indicators for spatial association analysis, \emph{e.g.}, Geary's C~\cite{geary_1954_contiguity}, Moran's I~\cite{moran_I}, etc.
Among them, Moran's I is perhaps the most widely used metric, which can be measured as:

\begin{equation}
I = \frac{n}{\sum_i\sum_j w_{ij}} \frac{\sum_i \sum_j w_{ij} (x_g^i - \overline{x}_g^{ij}) (x_g^j - \overline{x}_g^{ij})}{\sum_i(x_g^i - \overline{x}_g^{ij})^2}
\end{equation}

\noindent
where $x_g^i$ indicates traffic volume of grid $i$ surrounding grid $g$, $\overline{x}_g^{ij}$ is the mean volume of all surrounding grids, $n$ is the number of spatial grids indexed by $i$ and $j$, and $w_{ij}$ is an element of spatial weights.
Here, we opt to the first-order queen contiguity spatial weight matrix, which is a common choice in the literature~\cite{zhang_2017_quantifying, nelson_2017_evaluating}.
We choose $3\times3$ for the matrix size, corresponding to the setting of convolution kernel size adopted by ST-ResNet.

To support the investigation of each region, we decompose the global Moran's I into local indicators of spatial association (LISA)~\cite{anselin_1995_lisa} indices.
In this way, each region can be represented as a point in the scatterplot as shown in Fig.~\ref{fig:moran}.
The point position corresponds to traffic volume of region along x-axis, and traffic volume of local tract along y-axis.
The point color indicates prediction error.
To couple with multi-scale comparison, we standardize traffic volumes and prediction errors with a mean of 0 and variance of 1.
The mean of all LISA indices is proportion to the global Moran's I, which is represented as a regression line.
The correlation and confidence are also presented.
Selected points will be highlighted in blue and the point sizes are enlarged; see an example in Fig.~\ref{fig:moran}.

Fig.~\ref{fig:moran} presents the corresponding scatterplot for the bivariate map in Fig.~\ref{fig:bivariate_map}.
Most points are positioned around the regression line.
A correlation value of 0.793 and confidence $p < 0.01$ indicate a strong positive correlation. 
The point colors gradually change from \red{light yellow} (low prediction error) to dark yellow (high prediction error) from left to right, whilst marginal changes are observed in y-dimension.
This indicates that the prediction accuracy is more dependent on traffic volume of region, rather than traffic volume of local tract.



\subsection{Multi-scale Attribution View}
\label{ssec:unit_plot}
We design Multi-scale Attribution View to support scale-independent comparison (\textbf{T.3}).
To enable the comparison of individual regions, we opt to unit visualization techniques rather than an aggregated visualization presenting a summary statistic.
This is nevertheless a nontrivial task, because there are huge amounts of regions to display and the data attributions exhibit dynamic variances.
To this end we choose dot plots~\cite{wilkinson_1999_dot}, which encodes each data point as a dot.
However, conventional dot plots using constant dot size cannot effectively address scalability and dynamic variance issues.
Inspired by nonlinear dot plots~\cite{rodrigues_2018_nonlinear}, we employ adaptive dot sizes and propose a new layout algorithm to address these issues.

\begin{figure}[t]
  \centering
  \includegraphics[width=0.495\textwidth]{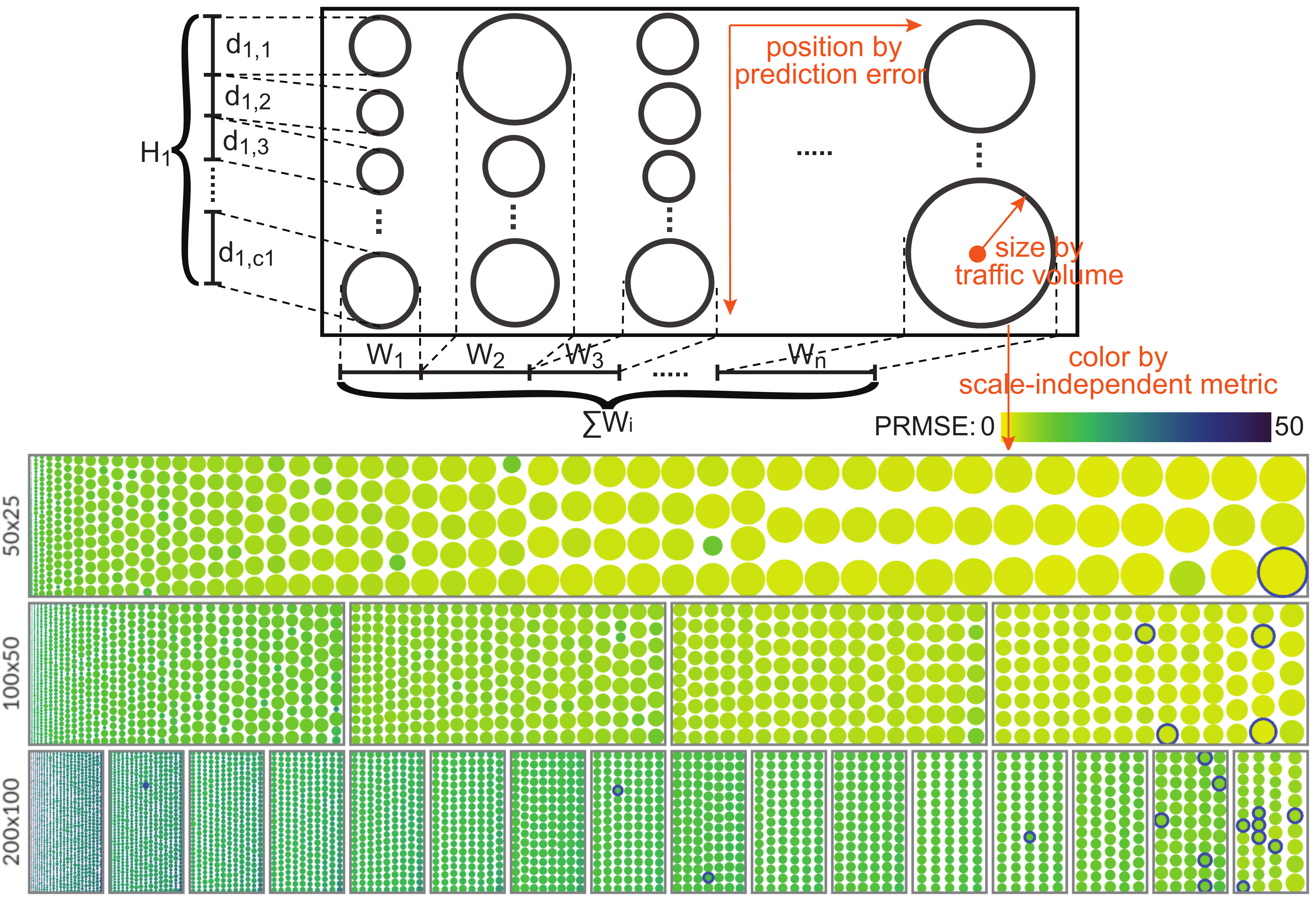}
  \vspace{-6mm}
  \caption{
  Multi-scale Attribution View: illustration of the layout algorithm for positioning dots in an enclosing rectangle (top);
  arrangement of three-scale dot plots in a hierarchy structure to facilitate comparison over multiple scales (bottom).
  \red{Highlighting a region at a coarse scale will also highlight its sub-regions at finer scales.}
  }
  \vspace{-4mm}
  \label{fig:attribution}
\end{figure}

The plot is constructed through the following procedures:

\begin{enumerate}
\vspace{-1.5mm}
\item
\textit{Sorting}:
We first put all region into a list, which is the simplest and most common way to construct explanation~\cite{wang_2019_designing}.
The list is then sorted in ascending order by \red{absolute} prediction error, such that the region with high prediction errors will be emphasized.

\vspace{-1.5mm}
\item
\textit{Layout}:
Next, we place the regions in the display space, as illustrated in Fig.~\ref{fig:attribution} (top).
The layout algorithm takes input of an ordered list of data points $\mathcal{D} := \{D_1,\cdots,D_k\}$ where $D_i$ indicates a region, together with an enclosing rectangle $(W, H)$ where $W$ \& $H$ indicate the width and height, respectively.
The layout problem is essentially to divide $\mathcal{D}$ into $n$ columns $\mathcal{C} := \{C_1,\cdots,C_n\}$, where each column $C_i$ contains a set of data points $\{D_{i,1},\cdots,D_{i,c_i}\} \subseteq \mathcal{D}$.

For each $D_{i,j}$, we calculate its diameter in proportion to traffic volume of the region, denoted as $d_{i,j}$;
and for each $C_i$, we can derive the width $W_i$ as $W_i = max(d_{i,1},\cdots,d_{i,c_i})$, and the height $H_i = \sum_{j}^{c_i} d_{i,j}$.
We formalize the problem as a constrained optimization problem, with an objective to find the optimal column number $n$ and row number $c_i$ of each column $C_i$, such that the height of each column approximates the average height of all columns $\overline{H} = \sum_{i=1}^{n}H_i / n$, and the aspect ratio is as close as possible to that of the enclosing rectangle:

\vspace{-2.5mm}
\begin{equation}
\underset{n,c_i}{\arg\min} \sum_i |\sum_{j=1}^{c_i}d_{i,j}- \overline{H}| + |\frac{\sum_i W_i}{\overline{H}} - \frac{W}{H}|
\end{equation}

\vspace{-1.5mm}
constrained to $\sum_{i=1}^n c_i = k \, , 0< n,c_i < k$.
We solve the problem by firstly initializing $n$ with an estimated value $n = \sqrt{k \times \frac{W}{H}}$, and $c_i$ to satisfy $\sum_{j=1}^{c_i}d_{i,j} \approx \sum_{i=1}^{k}\frac{d_i}{n}$.
The optimal variable values are achieved through some turbulence.

\vspace{-1.5mm}
\item
\textit{Color coding}:
We color code each dot according to a scale-independent metric (see Sec.~\ref{ssec:statistics}) specified by users.
For example, dots in Fig.~\ref{fig:attribution} (bottom) are colored by \emph{PRMSE}.

\end{enumerate}

\noindent
\textit{Arrangement}:
To facilitate multi-scale comparison, we organize three-scale dot plots under the same partition shape in a hierarchical structure, as seen in Fig.~\ref{fig:attribution} (bottom).
Specifically, we divide the data points into four subsets at scale 100$\times$50, where each subset posses around one-fourth of all traffic volumes; then we generate one dot plot for each subset, and arrange the four dot plots side-by-side.
Similarly, we generate 16 dot plots at scale 200$\times$100.
This arrangement reminds users that a partition at scale 50$\times$25 corresponds to four partitions at scale 100$\times$50, and 16 partitions at scale 200$\times$100.

In summary, the Multi-scale Attribution View utilizes the following visual channels to represent data attributions.

\begin{itemize}

\vspace{-1.5mm}
\item
\emph{Position} encodes \red{absolute} prediction error.
At each scale, dots are positioned from left to right in ascending order of prediction errors;
within each column, the dots are positioned from top to bottom in ascending order of prediction errors.

\vspace{-2.5mm}
\item
\emph{Size} encodes traffic volume.
Since total traffic volumes at all three scales are the same, dot sizes are comparable across scales.

\vspace{-2.5mm} 
\item
\emph{Color} encodes one of the scale-independent evaluation metrics.
The three-scale dot plots share the same colormap, thus the dot colors are also comparable across scales.

\end{itemize}

\begin{figure}[tb]
  \centering
  \includegraphics[width=0.475\textwidth]{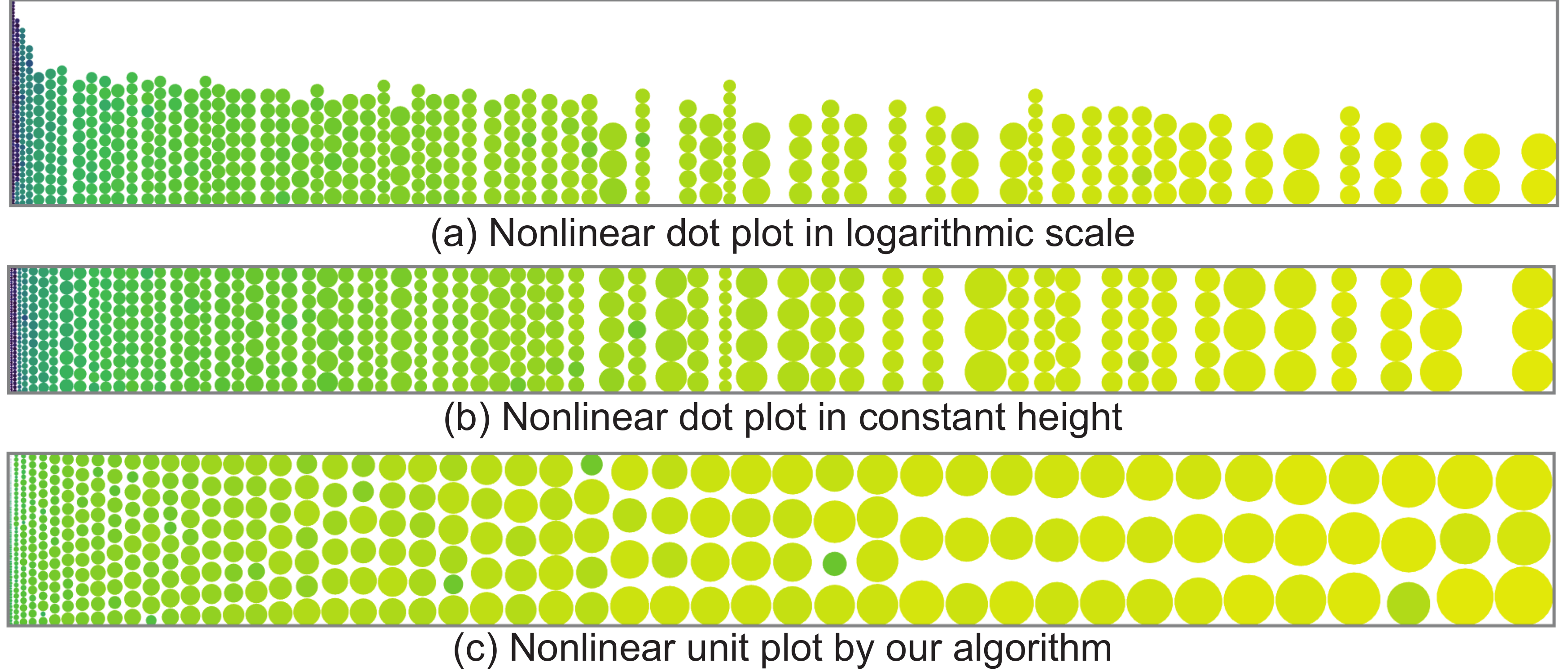}
  \vspace{-4mm}
  \caption{Alternative designs: nonlinear dot plots in logarithmic scale (a) and constant height (b) by~\cite{rodrigues_2018_nonlinear}, and by our algorithm (c).}
  \vspace{-5mm}
  \label{fig:dotplot_comp} 
\end{figure}

\vspace{-1.5mm}
\noindent
\textbf{Alternative design.}
An alternative design here is nonlinear dot plots~\cite{rodrigues_2018_nonlinear}.
The paper proposed many strategies for adapting dot sizes, such as making the column height following logarithmic scale (Fig.~\ref{fig:dotplot_comp}(a)) or constant height (Fig.~\ref{fig:dotplot_comp}(b)).
In comparison to the plot by our algorithm (Fig.~\ref{fig:dotplot_comp}(c)), those size adaption strategies produce different visual representations of data attributions, as follows:

\begin{itemize}
\vspace{-1.5mm}
\item
\textit{Position}:
In Fig.~\ref{fig:dotplot_comp}(a\&b), each stack is positioned horizontally based on prediction error of the dot in the bottom.
The other dots in the stack are those data points nearby.
In this sense, positions of most dots infer only relative ordering, as the same with ours.
However, the stacking strategy will leave many paddings in-between stacks, causing space usage deficiency. 

\vspace{-1.5mm}
\item
\textit{Size}:
Dot sizes in Fig.~\ref{fig:dotplot_comp}(a\&b) are adjusted based on the number of dots in each stack, which is essentially the frequency of data distribution.
In contrast, our algorithm determines dot size according to explicit traffic volume, which promotes correlation analysis between input traffic and output prediction.
As shown in Fig.~\ref{fig:dotplot_comp}(c), dot size generally grows from left to right.

\vspace{-1.5mm}
\item
\textit{Color}:
All plots encode scale-independent metrics using color.
However, their algorithm may cause misleading correlation analysis, and spoil multi-scale comparison because dot sizes reflect data distributions instead of explicit traffic volume.
As shown in Fig.~\ref{fig:dotplot_comp}(a\&b), the left part exhibits obvious \red{plum} colors.
Users may perceive high \emph{PRMSE} values, however these regions count up to insignificant traffic volume as in Fig.~\ref{fig:dotplot_comp}(c).
\end{itemize}


\subsection{User Interactions}
\label{ssec:interactions}

\begin{figure}[tb]
  \centering
  \includegraphics[width=0.499\textwidth]{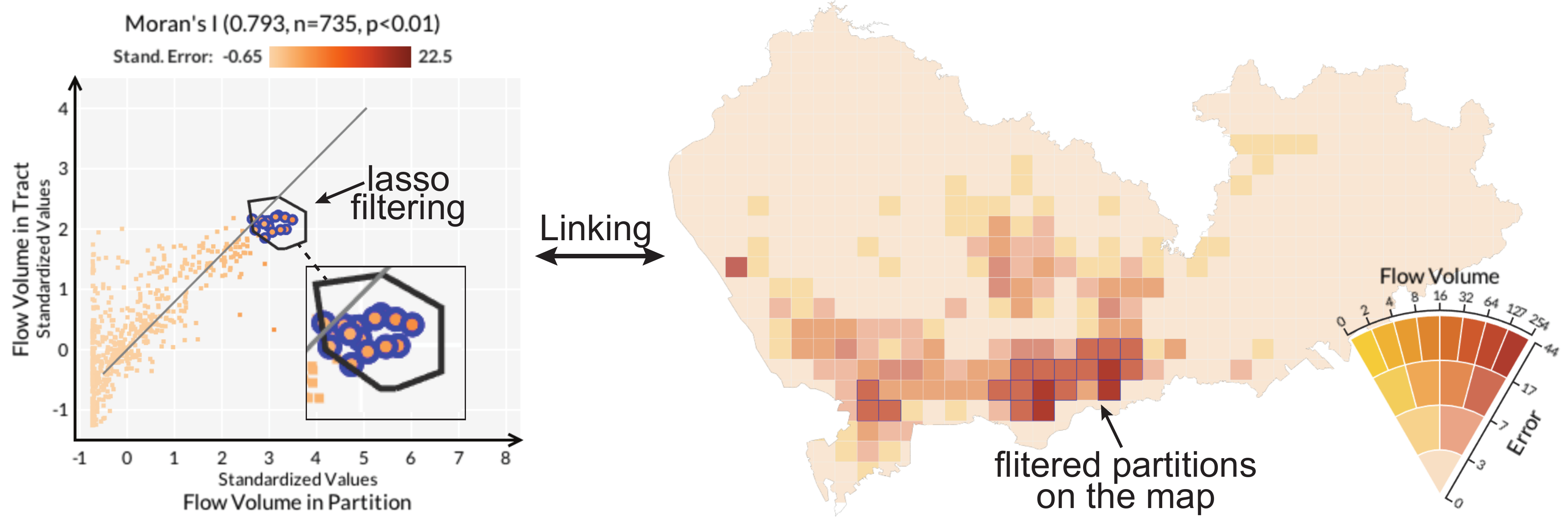}
  \vspace{-6mm}
  \caption{The views are linked: filtering points in the scatterplot (left) highlights corresponding grids on the map view (right).}
  \vspace{-3mm}
  \label{fig:linking} 
\end{figure}

In addition to basic map navigations, our system also integrates various interactions that enable:

\begin{itemize}
\vspace{-1mm}
\item
\textbf{Exploration}:
First, the interface includes widgets to explore partition scales (50$\times$25, 100$\times$50, and 200$\times$100) and shapes (grid and TAZ).
Second, \emph{Multi-scale Attribution View} allows users to select one out of the three scale-independent metrics (\emph{PRMSE}, \emph{U}, and \emph{CORR}).
The metric range is determined by the minimum and maximum values across all scales.

\vspace{-1mm}
\item \textbf{Selection \& Filtering}:
Users can select a specific data point of interest with \emph{Point} selection tool, or filter a subset of data points with \emph{Rect} or \emph{Lasso} tools.
The tools apply to all views, \emph{i.e.}, bivariate map, scatterplot, and attribution view.
Selected/filtered data points are highlighted in blue color.
Specifically for data points selected by \emph{Point} tool, heatmaps (see Fig.~\ref{fig:bivariate_map}(b)) are shown on the map view, presenting detailed variations of flow volumes and prediction errors in each time slot over seven days.

\item \textbf{Linking.}
Automatic linking among the three visualization modules is supported for coordination across multiple views.
Fig.~\ref{fig:linking} presents an example.
Here, we filter points of high prediction errors in the scatterplot with \emph{Lasso} tool, and corresponding regions will be highlighted in the map.
The regions are located in the southern part of Shenzhen, which borders on Hong Kong and is more developed than other regions.

\end{itemize}{}

\section{Evaluation}

\label{sec:evaluation}

\begin{wrapfigure}{HR}{0.15\textwidth}
\raggedleft
\includegraphics[width=0.15\textwidth]{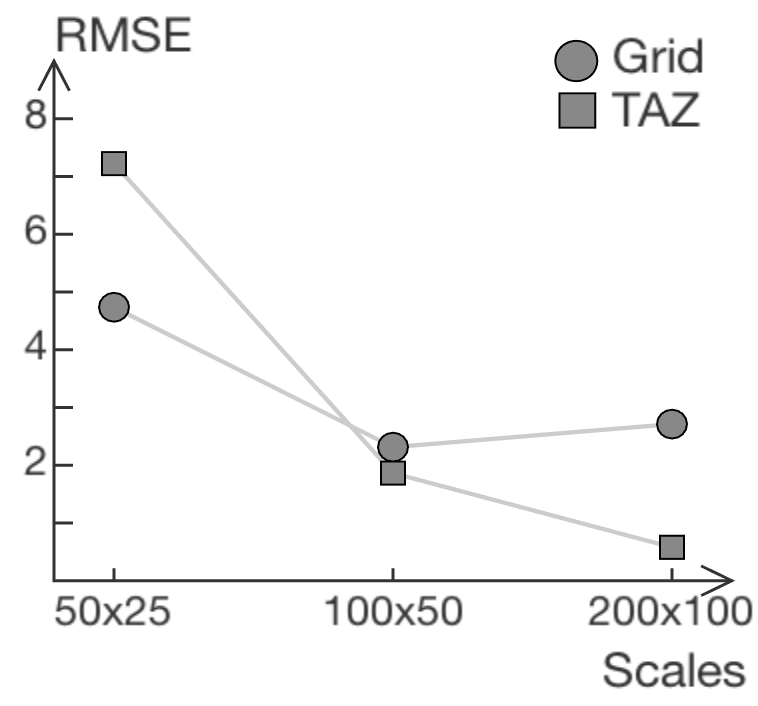}
\label{fig:rmse_inline}
\vspace{-6mm}
\end{wrapfigure}

The inline figure presents RMSEs generated by the six ST-ResNet models.
The statistic varies upon both partition shapes and scales.
For both Grid and TAZ partition, the coarsest scale 50$\times$25 yields the highest RMSE, and RMSE drops to similar values at scale 100$\times$50.
The decline could be due to either improvement in network predictions, or simply the increase in number of partitions.
Interestingly in the finest scale 200$\times$100, RMSE continues to drop for TAZ partition, but increases for Grid partition.

To unveil the underlying mechanism, we conduct three case studies in diagnosing predictions across multiple scales (Sec.~\ref{ssec:study1}) and by different partition shapes (Sec.~\ref{ssec:study2}), and exploring individual unit (Sec.~\ref{ssec:study3}).
In the end we present expert reviews (Sec.~\ref{ssec:review}). 

\subsection{Study 1: Diagnostics of Multiple-scale Predictions}
\label{ssec:study1}

From the above analysis, we observe that the RMSE varies upon partition scales.
To eliminate the confounding factor of partition number, this study compares predictions across multiple scales using scale-independent metrics.
Here, we select Grid partition and compare predictions of scales 50$\times$25, 100$\times$50, and 200 $\times$100.
Fig.~\ref{fig:teaser} presents the bivariate maps on the top, and the attribution view in the bottom.

All the bivariate maps present dynamic spatial variations:
dark colors (\emph{i.e.}, high prediction errors) are concentrated in the southern regions of the city, which are more developed areas and most taxi movements are there;
in contrast, the regions towards the north show light colors (\emph{i.e.}, low prediction errors).
Moreover, we can notice that Fig.~\ref{fig:teaser}(a)\&(b) share similar maximum prediction errors, but Fig.~\ref{fig:teaser}(b) presents less dark colors than Fig.~\ref{fig:teaser}(a).
This indicates that scale 100$\times$50 improves network predictions for many grids than scale 50$\times$25.
Instead, the maximum prediction error in Fig.~\ref{fig:teaser}(c) is two times of that in Fig.~\ref{fig:teaser}(b), whilst the two maps exhibit similar color distributions.
Hence the predictions are not improved from scale 100$\times$50 to scale 200$\times$100, which explains the increase of RMSE.

Fig.~\ref{fig:teaser}(d) compares uncertainty coefficient (U) across multiple scales.
We notice that dots at scales 50$\times$25 and 100$\times$50 are mostly in green or \red{lemon} (low uncertainties), whilst those at scale 200$\times$100 are mostly in \red{plum} (high uncertainties).
Specifically, we select two partitions from dot plot of scale 50$\times$25: partition 1 presents the highest prediction error but low uncertainty, while partition 2 shows a lower prediction error but high uncertainty.
The dots of their sub-partitions are also highlighted in the dot plots of scales 100$\times$50 and 200$\times$100.
We can observe that most sub-partitions of partition 1 have high prediction errors and low uncertainties, whilst most sub-partitions of partition 2 show low prediction errors and high uncertainties.


\begin{figure*}
  \centering
  \includegraphics[width=0.995\textwidth]{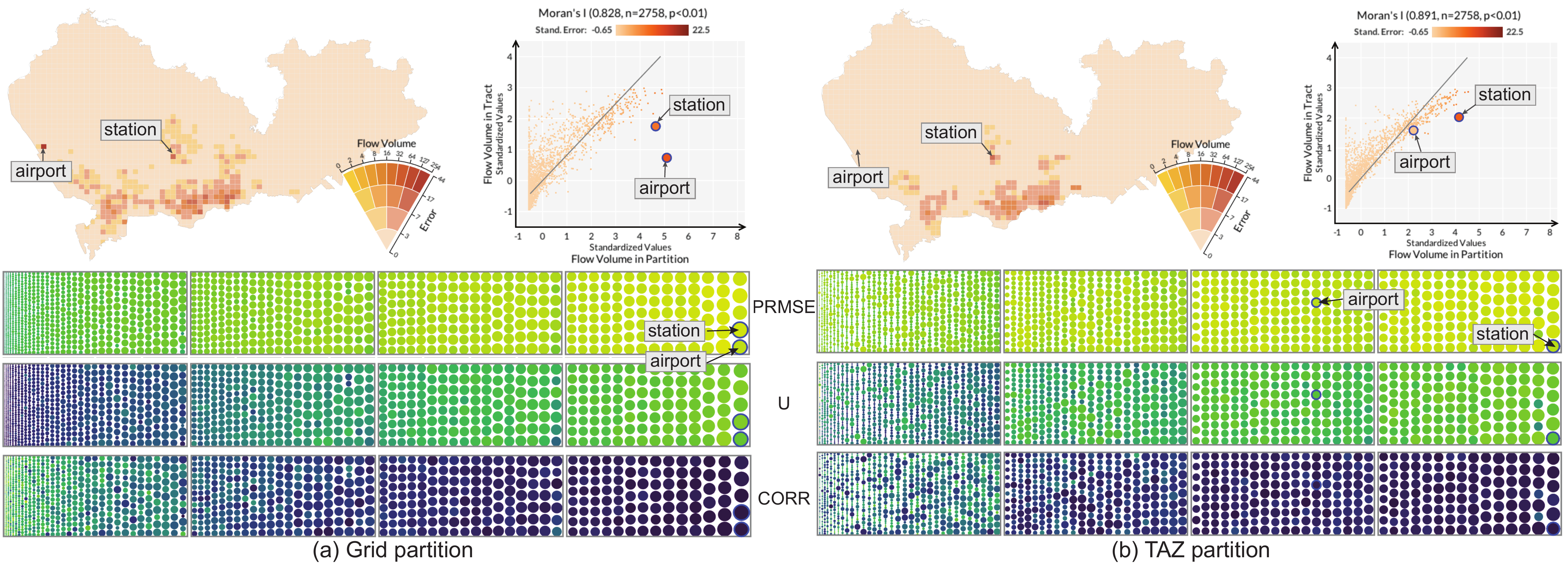}
  \vspace{-3mm}
  \caption{Comparing predictions by Grid partition (a) and TAZ partition (b) of the same scale 100$\times$50.
  Though the two partitions produce similar RMSEs, our system reveals that TAZ partition performs better in terms of \emph{PRMSE}, \emph{U}, and \emph{CORR}.}
  \vspace{-3mm}
  \label{fig:study2}
\end{figure*}

\subsection{Study 2: Comparison of Different Partition Shapes}
\label{ssec:study2}
Grid and TAZ partitions share similar RMSEs at scale 100$\times$50.
Nevertheless, RMSE is a summary statistic that can not reveal uniqueness of individual unit.
To overcome this limitation, this study compares predictions of individual units over different partition shapes.

Fig.~\ref{fig:study2} presents the results by Grid partition (a) and TAZ partition (b), both at scale 100$\times$50.
Overall the views present largely the same results, yet minor differences exist.
In the map views, neighboring regions share more similar predictions in Fig.~\ref{fig:study2}(b) than those in Fig.~\ref{fig:study2}(a).
This is expectable because rasterization in TAZ partition eventually smoothen traffic volumes in neighboring regions.
The scatterplots prove the argument, as the points are more concentrated nearby the regression line in Fig.~\ref{fig:study2}(b).
From the attribution views, we can observe that Fig.~\ref{fig:study2}(b) present more \red{lemon} dots in \emph{PRMSE}, especially for those dots on the left side.
That is, TAZ partition generates more accurate predictions than Grid partition in terms of \emph{PRMSE} at scale 100$\times$50, though their RMSEs are similar. 

In addition, individual units present rather different outputs.
Here we select two units with highest prediction errors in Grid partition: one locates the airport, while the other one locates a high-speed railway station. 
Both have high traffic volumes.
As shown in Fig.~\ref{fig:study2}(a), they are salient in the map view, and are far away from the regression line in the scatterplot.
On the other hand, the units are much less noticeable in Fig.~\ref{fig:study2}(b), especially the airport.
The collaborating expert \emph{CR} found a possible reason: the airport is located in a remote area where the TAZ is large $-$ traffic is shared with neighboring regions; in contrast, the station is in the city center where the TAZ is small $-$ traffic is not shared with neighboring regions.

\begin{figure}
  \centering
  \includegraphics[width=0.495\textwidth]{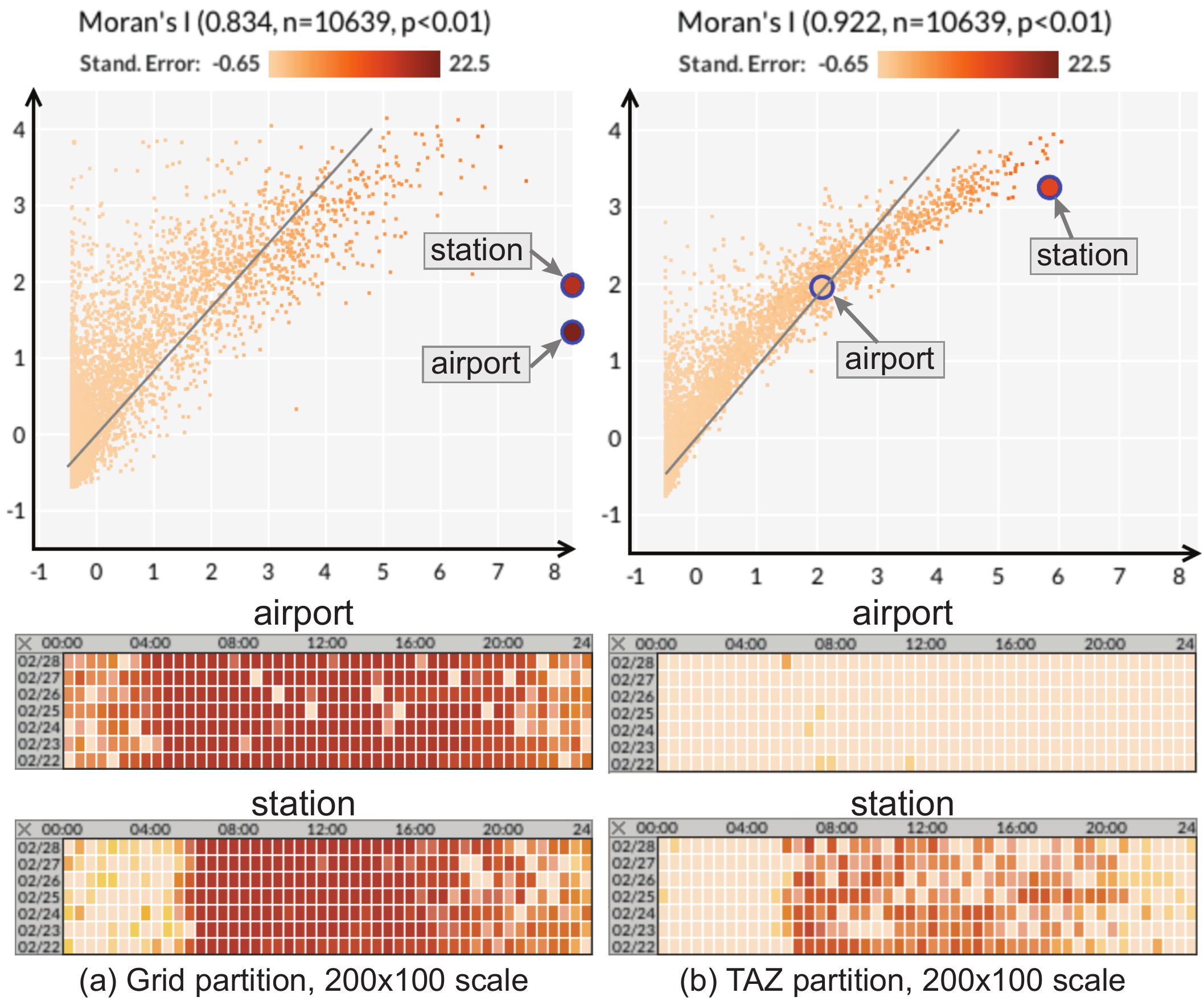}
  \vspace{-4mm}
  \caption{Investigating prediction variations over time of individual regions. The airport shows great differences in Grid partition (a) and TAZ partition (b), while the difference is minor for the station.}
  \vspace{-4mm}
  \label{fig:study3}
\end{figure}

\subsection{Study 3: Investigation of Individual Units}
\label{ssec:study3}
To understand why RMSE of TAZ partition at scale 200$\times$100 continues to decline while that of Grid partition increases, this study conducts in-depth investigation of prediction variations over time of individual units.
Fig.~\ref{fig:study3} presents temporal views of the airport and station by Grid partition (a) and TAZ partition (b) at scale 200$\times$100.
By comparing the scatterplots in Fig.~\ref{fig:study3} with those in Fig.~\ref{fig:study2}, we can observe that the points in Fig.~\ref{fig:study3}(a) (Moran's I 0.834) are more sparse than those in Fig.~\ref{fig:study2}(a) (Moran's I 0.828) $-$ spatial heterogeneity increases when Grid partition scales up.
On the other hand, the points in Fig.~\ref{fig:study3}(b) (Moran's I 0.922) are more concentrated that those in Fig.~\ref{fig:study3}(b) (Moran's I 0.891) $-$ spatial heterogeneity remains when TAZ partition scales up.

In Fig.~\ref{fig:study3}(a), the airport and station are far away from the regression line, and the points are in \red{dark orange} colors.
Their partition/tract volume ratios are bigger, indicating more granularity in Grid partition enlarges traffic volume differences between neighboring partitions.
By referring to their temporal views, we can notice that the airport exhibits high traffic volumes and high prediction errors throughout the whole day, whilst the station shows low prediction errors between midnight to six o'clock in the early morning.
The difference is likely due to the fact that high-speed railway service are terminated before dawn, while the flight service operates all day.
From the comparison, we can observe that the airport and station points are much closer to the regression line in Fig.~\ref{fig:study3}(b).
The station point in Fig.~\ref{fig:study3}(b) is still in \red{orange, but lighter than that in Fig.~\ref{fig:study3}(a)}.
The airport point changes to \red{light orange}, indicating a lower prediction error.
The differences are more visible in the temporal views, where the airport unit exhibits mostly light colors. 

\subsection{Expert Review}
\label{ssec:review}

We conducted interviews with two independent experts (denoted as \emph{EA} and \emph{EB}) other than our collaborating researcher \emph{CR}.
Both experts are specialized in transportation, and have been actively working on traffic management for several years.
Each interview lasted for around one hour. 
In the first thirty minutes, we explained visual designs adopted in the system, demonstrated how the system works, and presented case studies. 
Next, we allowed them to explore the system for about twenty minutes.
In the end, we collected their feedbacks.

\vspace{1mm}
\noindent
\emph{Methodology.}
Both experts 
have experimented with deep learning models, but ``most often the outcomes are suspicious".
\emph{EB} pointed out ``interpretable outcomes will make deep learning more useful in traffic management". 
In this sense, both experts appreciated the efforts on developing a visual analytics to diagnose the predictions.
They also agreed with \emph{CR} to start with the MAUP, which is a hot topic in transportation and geography.
They especially appreciated the capability of investigating an individual region, which was not supported by most works they followed up.

\vspace{1mm}
\noindent
\emph{Interactive Visual Design.}
Both experts confirmed that the interface is nicely designed in accordance with the analytical tasks.
They agreed with the choice of multiple views to depict information from multiple perspectives, and they appreciated the linking among views.
\emph{EA} highlighted ``it is important that I can select a partition in the scatterplot or attribute view, and see where it is on the map".
All experts (including \emph{CR}) were not aware of the term `unit visualization', though they have utilized choropleth map and scatterplot before.
They easily adapted to the concept and felt ``definitely better than summary statistics (\emph{e.g.}, the RMSE figure) that omit details".
\emph{EA} and \emph{EB} were familiar with conventional bivariate colormaps provided in GIS software such as Esri ArcGIS, but not with the VSUP~\cite{michael_2018_value-suppressing}.
After understanding the visual encodings, they acknowledged that VSUP is more suitable in this work, as VSUP leverages fewer colors and emphasizes partitions with higher prediction errors. 
\red{In addition, all experts (including \emph{CR}) agreed that the bivariate map using VSUP surpasses the performance of side-by-side maps.} From Fig.~\ref{fig:bivariate_map}, ``we know which partitions should be examined", \emph{EA} commented.

The experts felt some difficulty in understanding the multi-scale attribution view.
They fully comprehended the visual encodings only after we illustrated how the view is constructed (Fig.~\ref{fig:attribution}) and showed the comparison with non-linear dot plots (Fig.~\ref{fig:dotplot_comp}).
At first \emph{EA} showed his preference of the non-linear dot plot in constant height (Fig.~\ref{fig:dotplot_comp}(b)), which ``is easy to understand".
He nevertheless agreed that our design ``is more accurate and useful", after we explained that over-sized dots of low-volume partitions could cause misleading correlation analysis. 
Both experts liked the arrangement of multiple dot plots in an hierarchical structure, which ``facilitates the multi-scale analysis of spatial heterogeneity", \emph{EA} pointed out.

\vspace{1mm}
\noindent
\emph{Applicability.}
\red{
The experts were intrigued to find out how the MAUP affects deep traffic prediction.
As depicted in the Moran's I scatterplot (Fig.~\ref{fig:moran}), prediction errors increase in line with flow volumes in partition, but not with those in local tract.  
The difference is proved in study 2 that compares predictions of the airport and the highway station.
Here peak flow volumes of the airport are distributed to the neighboring partitions under TAZ partition, and the prediction error drops dramatically.}
The studies demonstrated that TAZ partition better supports deep traffic prediction, coinciding with many empirical studies in traffic analysis relying on spatial partition.
Transportation researchers have gained much experience in finding proper spatial partitions, and the studies illustrate how the experience could be applied to improve deep learning models.
Based on the insights, the experts can ``focus on generating reasonable input features, rather than tuning parameters of complex neural networks that we are not familiar with", \emph{EA} suggested.

\section{Discussion}
\label{sec:discussion}

The studies provide several illuminating insights:
Study 1 shows that finer grained partition scales may generate worse scale-independent metrics.
The result is opposite with that \red{derived} from RMSE (the inline figure in Sec.~\ref{sec:evaluation}) $-$ a performance metric widely adopted for deep traffic prediction.
\red{Taking Grid partition for an example, RMSE suggests that scale $100\times50$ achieves better performance than the other two scales, but Fig.~\ref{fig:attribution} reveals that the coarsest scale $50\times20$ produces lowest PRMSE as most dots in the scale are in lemon color.
A possible cause for this phenomena is that prediction errors are linearly correlated with flow volumes, and scale-independent metrics can cope with the correlations but not RMSE.}
Hereby, we suggest that scale-independent metrics should be used in future studies for fair comparison across different scales. \red{Besides, a promising direction for improving the prediction performance is to introduce attention mechanism~\cite{vaswani_2017_attention} that emphasizes regions with high flow volumes.}

Study 2 reveals that TAZ partition is more suitable than Grid partition for deep traffic prediction.
And study 3 depicts a probable reason $-$ TAZ partition reduces the number of outliers by averaging peak traffic into neighboring regions.
\red{From average values of PRMSE, CORR, and U, we notice that PRMSE and uncertainty increase while correlation drops, indicating prediction performance drops, when scaling up for Grid partition. 
In contrast, PRMSE and uncertainty decrease while correlation increases, indicating prediction performance improves, when scaling up for TAZ partition.
This is probably because spatial heterogeneity decreases when scaling up for TAZ partition, while that increases when scaling up for Grid partition, and deep learning models can better predict traffic of regions with low spatial heterogeneity.}
The finding leads to promising directions on how to improve traffic prediction without tedious hyperparameter tuning in neural networks.
For instance, in addition to Grid and TAZ, geographical partitions can also be formulated based on human activities through spatial clustering~\cite{andrienko_spatial_2011-1, wu_2017_mobiseg}, or graph partitioning~\cite{guo_2009_flow, wang_2019_visualizing}. 
Besides, gradual partition~\cite{Moeckel_2015_gradual} could generate more balanced traffic partitions, which could also bring a positive effect on traffic prediction.

The insights are disclosed with unit visualization techniques, which are in high demand by GIS and transportation colleagues~\cite{nelson_2017_evaluating}.
These visualizations however also face significant scalability issue.
For example, the scatterplots (Fig.~\ref{fig:study3}) of scale 200$\times$100 suffer from some amount of overplotting.
If finer partitions are employed (\emph{e.g.}, scale 400$\times$200), the occlusion would be even worse, and dots in the non-linear dot plot become tiny.
Proper aggregations (\emph{e.g.},~\cite{goodwin_2016_visualizing, zhang_2016_visual}) can be incorporated to mitigate such issues.

\vspace{1mm}
\noindent
\emph{Limitation and Future Work.}
The current system exhibits some limitations.
First, this work only examined grid and TAZ partition shapes, whilst many other partition units are available.
For instance, segmenting the territory based on road network could reflect urban traffic better.
The experts expected the road-based partition could produce more accurate traffic predictions.
We would like to examine these partition methods in the future work.
Second, the system employs inconsistent colormaps among the three visualization modules.
For example, the sequential colormap in the scatterplot is different with the 4-class OrRd colormap in the bivariate map, even though both colors encode prediction errors.
In fact, we experimented with the settings, but unsatisfactory visualizations were generated.
In addition, prediction errors in the scatterplot are standardized, while those in the bivariate map are absolute values $-$ they are not the same.
Hence, we opt to different colormaps in the end.
Nevertheless, it is worthwhile to strive for consistency in the visual encodings across the different views to support the visual analysis further.
\red{Finally, ST-ResNet model adopted in this work applies 2D convolution operations on temporal-varying matrices.
The model requires to partition the entire region into non-overlapping grids, which however casts away neighborhood relationships between TAZs and may cause negative effect on prediction performance.
The deficiency can be overcome using graph convolutional networks (GCN) with graph convolution operations.
We would like to employ GCN that can encapsulate both spatial and temporal attributes at the same time to model traffic predictions.}

\section{Conclusion}
\label{sec:conclusion}

This paper presents a visual analytics approach for diagnosing the MAUP in deep traffic prediction.
Through discussions with a collaborating expert, we identify various analysis criteria and formulate a set of analytical tasks.
To cope with the analytics tasks, we 
(i) train six ST-ResNet~\cite{zhang_2017_deep} models by applying Grid and TAZ partition shapes, and three partition scales of 50$\times$25, 100$\times$50, and 200$\times$100, to the underlying studying area;
(ii) employ scale-independent metrics, instead of the conventional RMSE, to evaluate network predictions; and
(iii) develop a visual analytics system integrating three visualization modules, namely \emph{Bivariate Map}, \emph{Moran's I Scatterplot}, and \emph{Multiscale Attribution View}.
All the views adopt unit visualization techniques that support investigation on a single data point.
Specifically, we employ a value-suppressing uncertainty palette~\cite{michael_2018_value-suppressing} in the bivariate map, and we design a new layout strategy for nonlinear dot plots, which is more space efficient and trustworthy than existing layout methods, in the multiscale attribution view.
The designs are well recognized by transportation experts.
In the end, we conduct three case studies on a real-world taxi data in Shenzhen, which reveal several insightful findings.
For example, 
predictions can be improved by more evenly distributed traffic aggregations.
Feedbacks from independent experts also confirm the effectiveness of our system.

\vspace*{1.5mm}
\noindent
\acknowledgments{
The authors wish to thank the independent experts and the anonymous reviewers for their valuable comments.
This work is supported in part by National Natural Science Foundation of China (61802388, 61702433, 61872389, 41701452).
Wei Chen is supported by National Natural Science Foundation of China (61772456, U1609217).
}

\bibliographystyle{abbrv}
\bibliography{Reference}

\end{document}